\documentclass[journal]{IEEEtran}
\usepackage{amsmath,amsthm,amsfonts,amssymb,bm}
\usepackage{algorithm}
\usepackage{algpseudocode}
\usepackage{array}
\usepackage[caption=false,font=normalsize,labelfont=sf,textfont=sf]{subfig}
\usepackage{textcomp}
\usepackage{stfloats}
\usepackage{url}
\usepackage{verbatim}
\usepackage{graphicx}
\usepackage{cite}

\usepackage{booktabs}
\usepackage{threeparttable}
\usepackage{multirow}
\usepackage{multicol}

\usepackage[table]{xcolor}
\usepackage[pdftex, hidelinks]{hyperref}

\graphicspath{{fig/}}

\usepackage{environ} 

\NewEnviron{change_env}{\textcolor{blue}{\BODY}} 

\algnewcommand{\LineComment}[1]{\Statex \# #1}

\begin{document}

\title{Rethinking Decoupled Knowledge Distillation: \\A Predictive Distribution Perspective}

\author{
  Bowen~Zheng and Ran~Cheng,~\IEEEmembership{Senior Member,~IEEE}
  \thanks{
    Bowen Zheng is with the Department of Data Science and Artificial Intelligence, The Hong Kong Polytechnic University, Hong Kong SAR, China. E-mail: \protect{bowen.zheng@protonmail.com}.
  }
  \thanks{Ran Cheng is with the Department of Data Science and Artificial Intelligence, and the Department of Computing, The Hong Kong Polytechnic University, Hong Kong SAR, China. He is also with The Hong Kong Polytechnic University Shenzhen Research Institute, Shenzhen, China. E-mail: ranchengcn@gmail.com.}
}

\markboth{Journal of \LaTeX\ Class Files,~Vol.~14, No.~8, August~2021}%
{Shell \MakeLowercase{\textit{et al.}}: A Sample Article Using IEEEtran.cls for IEEE Journals}


\maketitle

\begin{abstract}
  In the history of knowledge distillation, the focus has once shifted over time from logit-based to feature-based approaches.
  However, this transition has been revisited with the advent of Decoupled Knowledge Distillation (DKD), which re-emphasizes the importance of logit knowledge through advanced decoupling and weighting strategies.
  While DKD marks a significant advancement, its underlying mechanisms merit deeper exploration.
  As a response, we rethink DKD from a predictive distribution perspective.
  First, we introduce an enhanced version, the Generalized Decoupled Knowledge Distillation (GDKD) loss, which offers a more versatile method for decoupling logits.
  Then we pay particular attention to the teacher model's predictive distribution and its impact on the gradients of GDKD loss, uncovering two critical insights often overlooked: (1) the partitioning by the top logit considerably improves the interrelationship of non-top logits, and (2) amplifying the focus on the distillation loss of non-top logits enhances the knowledge extraction among them.
  Utilizing these insights, we further propose a streamlined GDKD algorithm with an efficient partition strategy to handle the multimodality of teacher models' predictive distribution.
  Our comprehensive experiments conducted on a variety of benchmarks, including CIFAR-100, ImageNet, Tiny-ImageNet, CUB-200-2011, and Cityscapes, demonstrate GDKD's superior performance over both the original DKD and other leading knowledge distillation methods.
  The code is available at \url{https://github.com/ZaberKo/GDKD}.
\end{abstract}


\section{Introduction}

\IEEEPARstart{O}{ver} the last decade, large neural networks have catalyzed major advancements in deep learning across various domains.
For instance, the convolutional neural networks have become a cornerstone in computer vision tasks such as image classification \cite{krizhevskyImageNetClassificationDeep2012}, \cite{heDeepResidualLearning2015}, \cite{zhangShuffleNetExtremelyEfficient2017,dingScalingYourKernels2022}, object detection \cite{renFasterRCNNRealTime2015}, \cite{liuSSDSingleShot2016}, and semantic segmentation \cite{ronnebergerUNetConvolutionalNetworks2015}, \cite{zhaoPyramidSceneParsing2017}.
Besides, the recent emergence of large-scale language and vision-language models \cite{openaiGPT4TechnicalReport2023,driessPaLMEEmbodiedMultimodal2023} has led to models with unprecedented sizes. Although these models benefit from large model capacity, they introduce substantial computational and storage challenges, particularly for mobile and embedded systems.
As a response, the Knowledge distillation (KD) \cite{hintonDistillingKnowledgeNeural2015} has become a critical model compression approach, which enables insights from larger models to be transferred to more compact counterparts.

KD's core concept involves leveraging the characteristic of overparameterization on larger models to enhance the generalization ability \cite{gouKnowledgeDistillationSurvey2020}, \cite{allen-zhuLearningGeneralizationOverparameterized2019}, \cite{aroraOptimizationDeepNetworks2018,brutzkusWhyLargerModels2019}.
In this procedure, a pre-trained \emph{teacher} model employs its advanced \emph{knowledge} to instruct a more streamlined \emph{student} model.
This transfer of knowledge is achieved through \emph{distillation}, a process where the complex intelligence of the teacher is condensed into essential insights, supervising the student to learn more efficiently.

Initially, Hinton \textit{et al.} \cite{hintonDistillingKnowledgeNeural2015} employed the teacher model's logits as the primary knowledge source in their logit-based knowledge distillation method, aiming to minimize the Kullback-Leibler (KL) Divergence between the soft softmax predictions of the teacher and student models.
This method, however, was later overshadowed by feature-based KD, as pioneered by FitNet \cite{romeroFitNetsHintsThin2015}, which emphasized the use of intermediate layer features of models for knowledge transfer.
The prevailing view suggested that feature-based knowledge offered better informational richness and was easier to assimilate than logit-based knowledge \cite{gouKnowledgeDistillationSurvey2020}, \cite{heoComprehensiveOverhaulFeature2019}.
Until very recently, the advent of Decoupled Knowledge Distillation (DKD) \cite{zhaoDecoupledKnowledgeDistillation2022} marked a significant paradigm shift, rekindling interest in logit-based methods.
Specifically, DKD revised the traditional KD loss by integrating the target label and leveraging the properties of KL-Divergence to decouple the distillation loss.
This subtle adjustment of logit decoupling enabled DKD to rival sophisticated feature-based methods, thus highlighting the efficacy of logit-based knowledge when distillation techniques are meticulously engineered.

\begin{figure*}[t]
  \centering
  \subfloat{
    \includegraphics[width=0.6\linewidth]{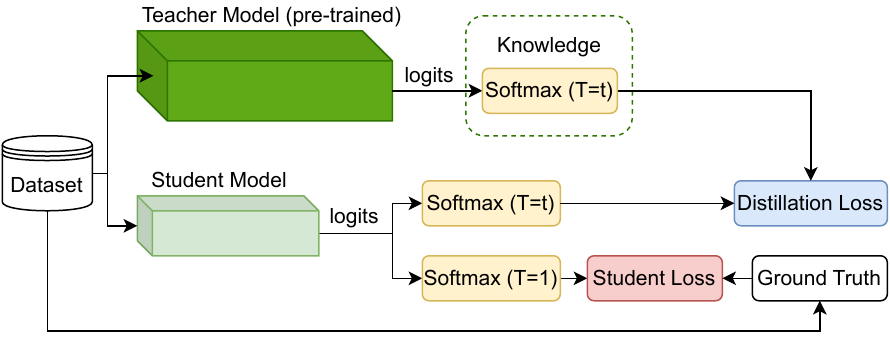}
  }
  \subfloat{
    \includegraphics[width=0.39\linewidth]{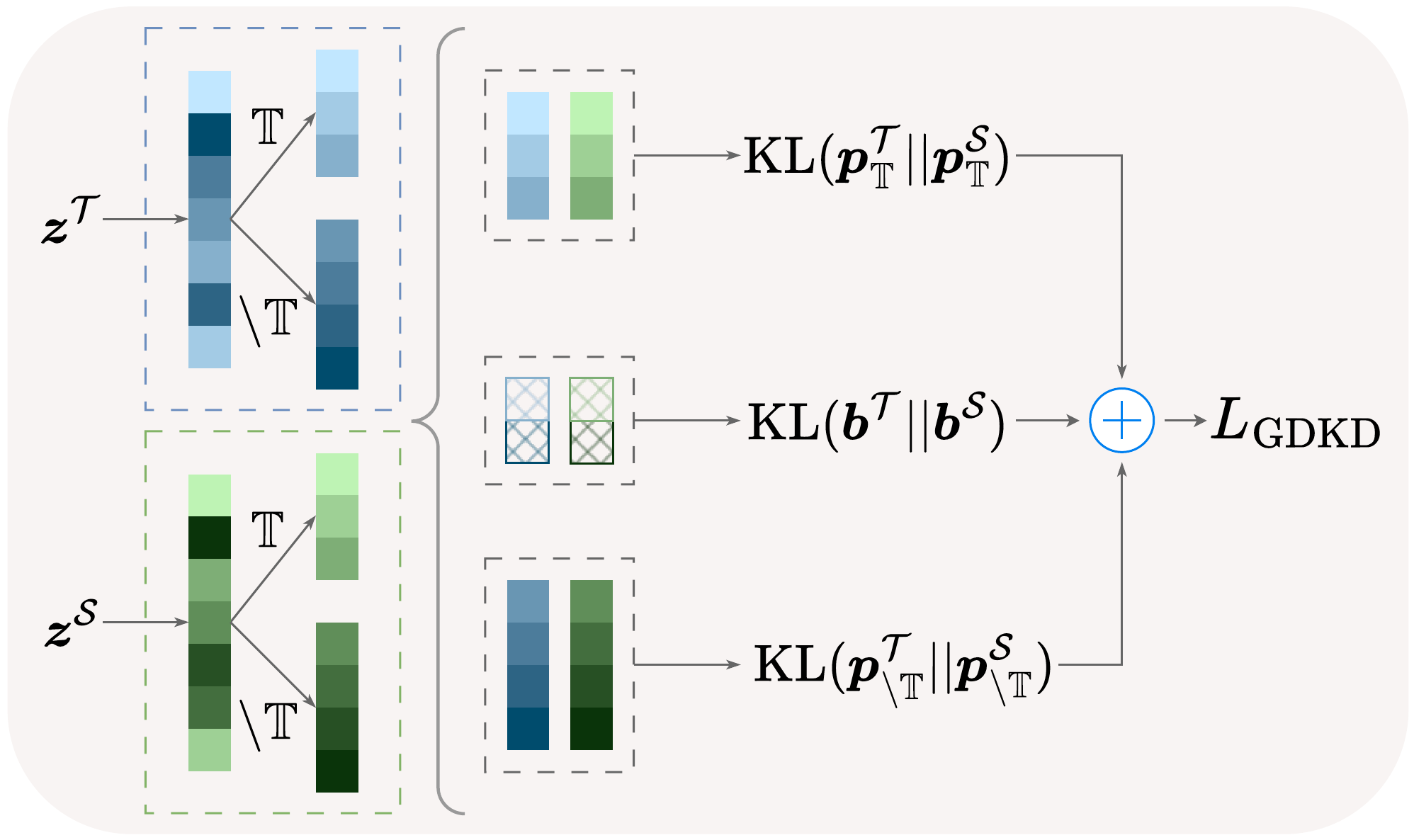}
  }
  \caption{\textbf{Left:} Logit-based knowledge distillation framework. The student model is supervised by the large pre-trained teacher's soft labels and target labels to enhance its performance. \textbf{Right:} Proposed GDKD loss under two-level decomposition with two partitions. Logits from the teacher $\mathcal{T}$ and student $\mathcal{S}$ are separately partitioned into two groups via class sets $\{\mathbb{T}, \setminus \mathbb{T}\}$, where $\mathbb{T}$ is a subset of class indices $\{1,\dots,C \}$.
    Then the distillation loss is calculated based on decoupled loss terms at the top-level binary distribution pair $(\bm{b}^\mathcal{T},\bm{b}^\mathcal{S})$ and the low-level distribution pairs $(\bm{p}_\mathbb{T}^\mathcal{T},\bm{p}_\mathbb{T}^\mathcal{S})$, $(\bm{p}_{\setminus \mathbb{T}}^\mathcal{T},\bm{p}_{\setminus \mathbb{T}}^\mathcal{S})$.}
  \label{fig:logits_kd}
\end{figure*}

While DKD has highlighted the value of logit-based knowledge in distillation, its intrinsic mechanisms have not been thoroughly explored.
To bridge the gap, our work begins by developing the \emph{Generalized Decoupled Knowledge Distillation (GDKD)} loss. As depicted in Fig.~\ref{fig:logits_kd}, GDKD loss advances the original KD loss by introducing a hierarchical partitioning of logits, thereby allowing more flexibility in the logit decoupling and including DKD as a specialized instance.

Then our investigation of the decoupling mechanism is conducted under this generalized framework, in the context of teacher models' predictive distributions. The empirical analysis reveals two crucial insights in terms of how decoupling patterns behind DKD can benefit from both top and non-top logits.
Capitalizing on these insights, we develop a streamlined GDKD algorithm, which efficiently processes multimodal predictive distributions (i.e., predictions with multiple peaks) and optimizes the utilization of logits. This algorithm offers a well-balanced performance in terms of accuracy and training speed. It inherits the advantages of logit-based knowledge distillation methods, necessitating no additional parameters compared to feature-based methods.

In summary, our contributions are:
\begin{itemize}
  \item \textbf{GDKD Loss}: We propose the GDKD loss, which refines and extends the concept of logit partitioning in knowledge distillation. GDKD offers a more versatile and effective approach compared to traditional methods, particularly in handling and representing non-top logits.

  \item \textbf{Empirical Insights}: Using the proposed GDKD loss, we perform a comprehensive empirical analysis of the underlying mechanisms of DKD. From the teacher model's predictive distribution perspective, we summarized two critical insights for logit-based knowledge distillation.

  \item \textbf{GDKD Algorithm}: Based on the above analysis, we propose a streamlined and efficient vanilla GDKD algorithm, which demonstrates efficacy in processing multimodal predictive distributions and enhances the learning capabilities of the student model.
        This algorithm, while simple, effectively balances model performance and training speed without requiring additional parameters, setting it apart from more complex feature-based methods.
\end{itemize}

Extensive experiments demonstrate GDKD's robust performance across a variety of benchmarks, especially in scenarios involving multimodal predictions: \textbf{(1)} In image classification benchmarks, GDKD consistently outperforms DKD, with accuracy improvements ranging from 0.22\%-1.12\% in CIFAR-100 and 0.57\%-1.20\% in ImageNet. \textbf{(2)} In transfer learning benchmarks, GDKD achieves highest accuracy in Tiny-ImageNet and CUB-200-2011; \textbf{(3)} In semantic segmentation benchmarks, GDKD records the highest mIoU scores on Cityscapes, surpassing other knowledge distillation methods in comparison.

\section{Related Work}
\label{sec:related_work}

Knowledge distillation, as conceptualized by Hinton \textit{et al.} in \cite{hintonDistillingKnowledgeNeural2015}, involves a large, pre-trained \emph{teacher} model $\mathcal{T}$ imparting knowledge to a smaller \emph{student} model $\mathcal{S}$.
Specifically, for \emph{logit-based knowledge distillation}, the distillation loss is based on the Kullback-Leibler (KL) divergence between the probability distributions of the teacher $\bm{p}^\mathcal{T}$ and the student $\bm{p}^\mathcal{S}$:
\begin{equation}
  \mathcal{L}_\text{KD} = \text{KL}(\bm{p}^\mathcal{T}||\bm{p}^\mathcal{S}),
  \label{eq:kd_loss}
\end{equation}
where $C$ is the number of classes and the probability distributions are calculated by the softmax of their logits outputs $ \bm{z} \in \mathbb{R}^C$ respectively:
\begin{equation}
  p_i = \frac{\exp(z_i)}{\sum_j^C \exp(z_j)}, i=1, \dots, C,
\end{equation}
with a temperature parameter $T$ incorporated to modulate the probability distribution via $z_i \gets z_i/T$.
On one hand, as $T$ tends towards infinity, the softmax-driven probability distribution becomes increasingly uniform, thus leading each class to have an equal probability of $1/C$.
On the other hand, when $T$ approaches zero, the distribution closely resembles one-hot labels, highlighting a more distinct classification.

While many logit-based distillation methods share a foundational framework, as depicted in Fig.~\ref{fig:logits_kd}, their distillation losses differ.
Soft labels from the teacher model, known as \emph{dark knowledge} \cite{hintonDistillingKnowledgeNeural2015}, play a crucial role in student learning via abstracted and condensed relations among classes.
However, the complexity and sufficiency of logit-based knowledge for student models have been questioned \cite{gouKnowledgeDistillationSurvey2020}, \cite{heoComprehensiveOverhaulFeature2019}.
Consequently, succeeding knowledge distillation techniques have employed intermediate features \cite{romeroFitNetsHintsThin2015}, \cite{zagoruykoPayingMoreAttention2022}, \cite{heoComprehensiveOverhaulFeature2019}, \cite{leeKnowledgeTransferDecomposing2022}, \cite{wangDistillingKnowledgeMimicking2022} and sample-level relations \cite{yimGiftKnowledgeDistillation2017}, \cite{parkRelationalKnowledgeDistillation2019}, \cite{tianContrastiveRepresentationDistillation2022}, \cite{yeGeneralizedKnowledgeDistillation2023}, sometimes alongside logits, to boost performance.

Nonetheless, recent studies \cite{zhaoDecoupledKnowledgeDistillation2022}, \cite{huangKnowledgeDistillationStronger2022}, \cite{zhaoGroupedKnowledgeDistillation2023}, \cite{jinMultiLevelLogitDistillation2023}, \cite{yangKnowledgeDistillationSelfKnowledge2023}, \cite{liCurriculumTemperatureKnowledge2023}, \cite{chiNormKDNormalizedLogits2023}, \cite{liuDCCDReducingNeural2024}, \cite{sunLogitStandardizationKnowledge2024} have revealed the untapped potential of logit-based knowledge distillation.
Recent logit-based distillation methods have further explored different facets of teacher-student relationships.
Methods such as DIST~\cite{huangKnowledgeDistillationStronger2022} and Multi-Level KD~\cite{jinMultiLevelLogitDistillation2023}
leverage the relaxed inter-class relations among logits as transferable knowledge.
Meanwhile, approaches like NCTK~\cite{liCurriculumTemperatureKnowledge2023} and LS~\cite{sunLogitStandardizationKnowledge2024}
introduce dynamic temperature schemes by applying different temperature settings to the teacher and student models.
In parallel, methods such as DKD~\cite{zhaoDecoupledKnowledgeDistillation2022} employ the classical soft-label paradigm through a decoupling framework,
highlighting the potential of logit partitioning in knowledge transfer.
In this work, we focus on the latter, aiming to enhance conventional logit distillation loss by dividing the logits based on the target label:
\begin{equation}
  \mathcal{L}_\text{KD} = \text{KL}(\bm{b}^\mathcal{T}||\bm{b}^\mathcal{S})+(1-p_t^\mathcal{T})\text{KL}(\bm{\hat{p}}^\mathcal{T}||\bm{\hat{p}}^\mathcal{S}),
  \label{eq:kd_loss_dkd}
\end{equation}
where $t$ is the true label of a given input, with $p_t^\mathcal{T}$ denoting the teacher's probability output for this true label.
In $\text{KL}(\bm{b}^\mathcal{T}||\bm{b}^\mathcal{S})$, the terms $\bm{b}^\mathcal{T}$ and $\bm{b}^\mathcal{S}$ represent the binary probability distributions of the target class for both the teacher and student models.
Meanwhile, in $\text{KL}(\bm{\hat{p}}^\mathcal{T}||\bm{\hat{p}}^\mathcal{S})$, $\bm{\hat{p}}^\mathcal{T}$ and $\bm{\hat{p}}^\mathcal{S}$ depict the distributions for non-target classes, recalculated using the softmax of non-target class logits.

Moreover, Zhao \textit{et al.} \cite{zhaoDecoupledKnowledgeDistillation2022} highlights that $\text{KL}(\bm{b}^\mathcal{T}||\bm{b}^\mathcal{S})$ provides insights on the difficulty of training samples, while $\text{KL}(\bm{\hat{p}}^\mathcal{T}||\bm{\hat{p}}^\mathcal{S})$ enhances the distillation performance.
And in traditional KD loss, $\text{KL}(\bm{\hat{p}}^\mathcal{T}||\bm{\hat{p}}^\mathcal{S})$ is coupled by the weight $1-p_t^\mathcal{T}$, thus suppressing the efficient knowledge transfer from it.
In response, DKD introduces a modified distillation loss:
\begin{equation}
  \mathcal{L}_\text{DKD} = \alpha \text{KL}(\bm{b}^\mathcal{T}||\bm{b}^\mathcal{S}) + \beta  \text{KL}(\bm{\hat{p}}^\mathcal{T}||\bm{\hat{p}}^\mathcal{S}).
  \label{eq:dkd_loss}
\end{equation}
Through judicious tuning of hyperparameters $\alpha$ and $\beta$, DKD facilitates the effective utilization of knowledge from both distillation loss terms.

\section{Methodology}
\label{sec:method}

In this section, we first revisit the logit-based knowledge distillation loss through a general two-level hierarchical decomposition and introduce the Generalized Decoupled Knowledge Distillation (GDKD) loss.
Then, we perform an empirical analysis to assess the impact of this decoupling from the teacher's predictive distribution.
By examining the gradients of logits under the GDKD loss during training, we identify two key factors that contribute to decoupled logit-based distillation. Building upon the empirical findings, we further propose an optimal logit partition strategy and the corresponding GDKD algorithm.

\subsection{GDKD Loss}
\label{sec:gdkd_loss}
Inspired by the foundation of the original DKD, we introduce the GDKD loss to further refine and expand the use of logit-based knowledge.
In GDKD, we conceptualize the class labels $\{1,\dots,C\}$ as being partitioned into two discrete sets: $\mathbb{T}$ and $\setminus \mathbb{T}$.
Within this framework, both the teacher and student models generate logits for each input.
Each logit output, denoted by $\bm{z} \in \mathbb{R}^C$, undergoes hierarchical decomposition into softmax predictions $\bm{p} \in \mathbb{R}^C$.
The top level of this decomposition is represented by $\bm{b}=[ b_\mathbb{T}, b_{\setminus \mathbb{T}} ]$, which encapsulates the aggregated binary probability distribution.
This is defined as $b_\mathbb{T}=\sum_{i \in \mathbb{T}} p_i$ and $b_{\setminus \mathbb{T}}=\sum_{i \in {\setminus \mathbb{T}}} p_i$, indicating the confidence level of each partition.
The leaf level consists of independent softmax probability distributions within these sets, $\bm{p}_\mathbb{T} \in \mathbb{R}^{|\mathbb{T}|}$ and $\bm{p}_{\setminus \mathbb{T}} \in \mathbb{R}^{|\setminus \mathbb{T}|}$:
\begin{gather}
  p_{i,\mathbb{T}} = \frac{\exp(z_i)}{\sum_{j \in \mathbb{T}} \exp(z_j) }, \\
  p_{i,\setminus \mathbb{T}} = \frac{\exp(z_i)}{\sum_{j \in \setminus \mathbb{T}} \exp(z_j) }.
\end{gather}
The KD loss, as reformulated in Eq.~\eqref{eq:kd_loss}, is the cumulative sum of these three KL-Divergence components:
\begin{equation}
  \begin{aligned}
    \mathcal{L}_\text{KD} = & \text{KL}(\bm{b}^\mathcal{T}||\bm{b}^\mathcal{S})                                                                                                                                                                                          \\
                            & + b_\mathbb{T}^\mathcal{T} \text{KL}(\bm{p}_\mathbb{T}^\mathcal{T}||\bm{p}_\mathbb{T}^\mathcal{S}) + b_{\setminus \mathbb{T}}^\mathcal{T} \text{KL}(\bm{p}_{\setminus \mathbb{T}}^\mathcal{T}||\bm{p}_{\setminus \mathbb{T}}^\mathcal{S}).
  \end{aligned}
  \label{eq:kd_loss_gdkd}
\end{equation}
\begin{proof}
  See Appendix~\ref{appendix:gdkd_loss_proof}
\end{proof}

Introducing novel weight factors $w_0, w_1, w_2$ leads to a decoupling of these components, culminating in the GDKD loss function:
\begin{equation}
  \begin{aligned}
    \mathcal{L}_\text{GDKD} = & w_0 \text{KL}(\bm{b}^\mathcal{T}||\bm{b}^\mathcal{S})                                                                                                                              \\
                              & + w_1 \text{KL}(\bm{p}_\mathbb{T}^\mathcal{T}||\bm{p}_\mathbb{T}^\mathcal{S})+w_2 \text{KL}(\bm{p}_{\setminus \mathbb{T}}^\mathcal{T}||\bm{p}_{\setminus \mathbb{T}}^\mathcal{S}).
  \end{aligned}
  \label{eq:gdkd_loss_3}
\end{equation}

\begin{figure}
  \centering
  \includegraphics[width=0.95\linewidth]{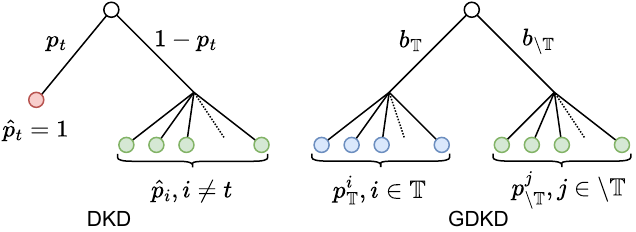}
  \caption{Illustration of logit partition strategies in DKD and GDKD. DKD utilizes a singular criterion, specifically the target label $t$, for logit partitioning. In contrast, GDKD employs a more flexible and sophisticated strategy, utilizing any mutually exclusive sets $\mathbb{T}$ and $\setminus \mathbb{T}$ for partitioning.
  }
  \label{fig:kd_decouple}
\end{figure}

\begin{figure*}[t]
  \centering
  \subfloat{
    \includegraphics[width=0.95\linewidth]{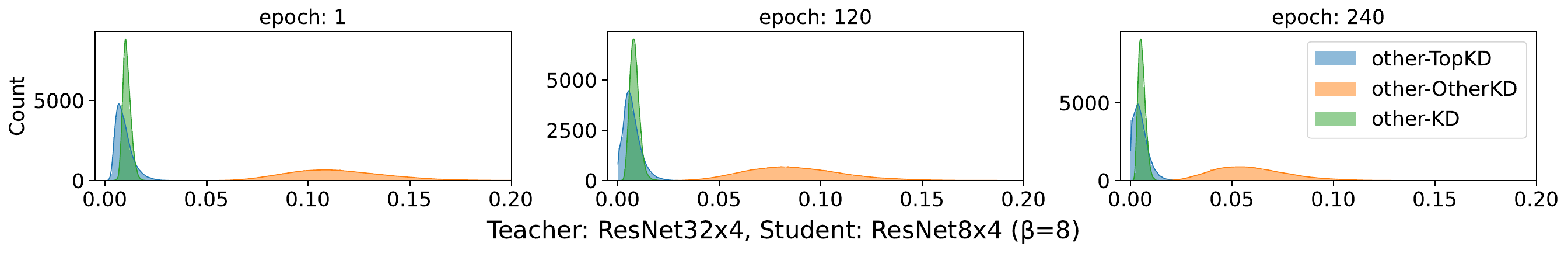}
  }\\
  \subfloat{
    \includegraphics[width=0.95\linewidth]{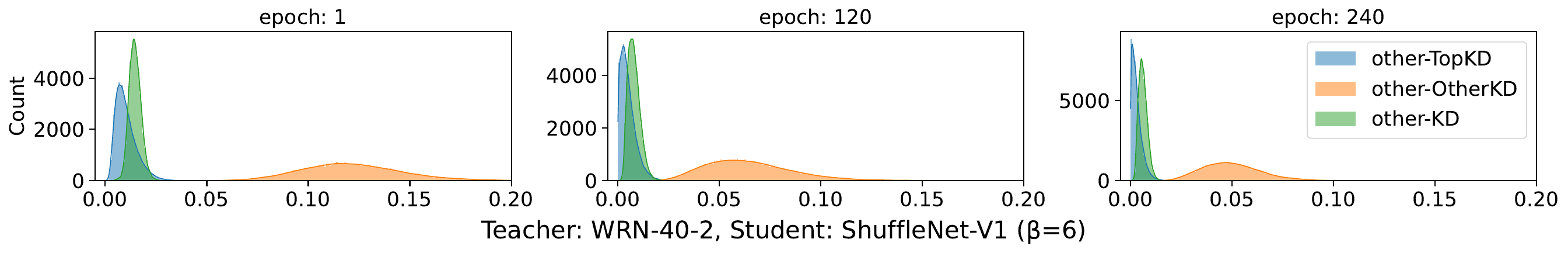}
  }
  \caption{Gradient magnitude distribution of non-top logits (other) in GDKD-top1 on CIFAR-100. The figure displays the distribution of non-top logits' average gradient magnitudes across training epochs under GDKD-top1 (where $\mathbb{T} = \{\arg\max_i (z_i^\mathcal{T})\}$). It includes distributions for $\mathcal{L}_\text{TopKD}$ (other-TopKD), $\beta\mathcal{L}_\text{OtherKD}$ (other-OtherKD), and the coupled term $(1-p_c^\mathcal{T}) \mathcal{L}_\text{OtherKD}$ (other-KD) from the traditional KD loss for comparison.}
  \label{fig:logits_grad_dist}
\end{figure*}

As depicted in Fig.~\ref{fig:kd_decouple}, GDKD's logit partition strategy is versatile.
If $\mathbb{T}$ only includes the target label ($\mathbb{T} = { t }$), the term $\text{KL}(\bm{p}_\mathbb{T}^\mathcal{T}||\bm{p}_\mathbb{T}^\mathcal{S})$ becomes null, thus effectively reducing Eq.~\eqref{eq:gdkd_loss_3} to the original DKD loss format.
This demonstrates that DKD is a specific instantiation of GDKD.
Furthermore, GDKD can be further extended to include more complex partitions by partitioning labels into multiple exclusive sets $\mathbb{T}_1,\dots,\mathbb{T}_n$, leading to a GDKD loss expression as follows:
\begin{equation}
  \mathcal{L}_\text{GDKD} =  w_0 \text{KL}(\bm{b}^\mathcal{T}||\bm{b}^\mathcal{S}) + \sum_{m=1}^n w_m \text{KL}(\bm{p}_{\mathbb{T}_m}^\mathcal{T}||\bm{p}_{\mathbb{T}_m}^\mathcal{S}).
  \label{eq:gdkd_loss_n}
\end{equation}
In addition, the partitioning can be recursively applied under multiple levels in the same manner. However, complex partitions introduce more hyperparameters for loss weights and risk diluting the original class relationships, potentially undermining the efficacy of logit-based knowledge distillation.
Therefore, this work primarily explores configurations of \emph{two-level decomposition with two partitions} to balance simplicity with preserving the information among logits.

\subsection{Empirical Analysis}
\label{sec:empirical_analysis}

In this study, GDKD is utilized as an analytical instrument to empirically explore the underpinnings of DKD. Initially, our focus is on a simplified variant of GDKD, termed $\text{GDKD}_2$. In this model, the set $\mathbb{T}$ encompasses only a single class $c$ from the complete label set $\{1, \dots, C\}$, with the rest of the classes falling into the complementary set $\setminus \mathbb{T}$:
\begin{equation}
  \mathcal{L}_{\text{GDKD}_2} = \text{KL}(\bm{b}^\mathcal{T}||\bm{b}^\mathcal{S}) + \beta \text{KL}(\bm{p}_{\setminus \mathbb{T}}^\mathcal{T}||\bm{p}_{\setminus \mathbb{T}}^\mathcal{S}) \footnote{In DKD \cite{zhaoDecoupledKnowledgeDistillation2022}, $\alpha$ in Eq.\eqref{eq:dkd_loss} is consistently set to 1. Therefore, for simplicity, the weights of $w_0$ in Eq.\eqref{eq:gdkd_loss_3} are also omitted in our discussion.}.
  \label{eq:gdkd_loss_2}
\end{equation}
In essence, DKD also represents a specific case of $\text{GDKD}_2$, with the chosen class $c$ matching the actual label $t$.

The empirical analysis is conducted on the CIFAR-100 dataset \cite{krizhevskyLearningMultipleLayers}, utilizing a ResNet32x4 as the teacher model and a ResNet8x4 as the student model \cite{heDeepResidualLearning2015}. The distillation temperature is fixed at $T=4$. Further details regarding the training specifics will be elaborated in Section~\ref{sec:exp}.

\begin{figure}
  \centering
  \subfloat[]{
    \includegraphics[width=0.95\linewidth]{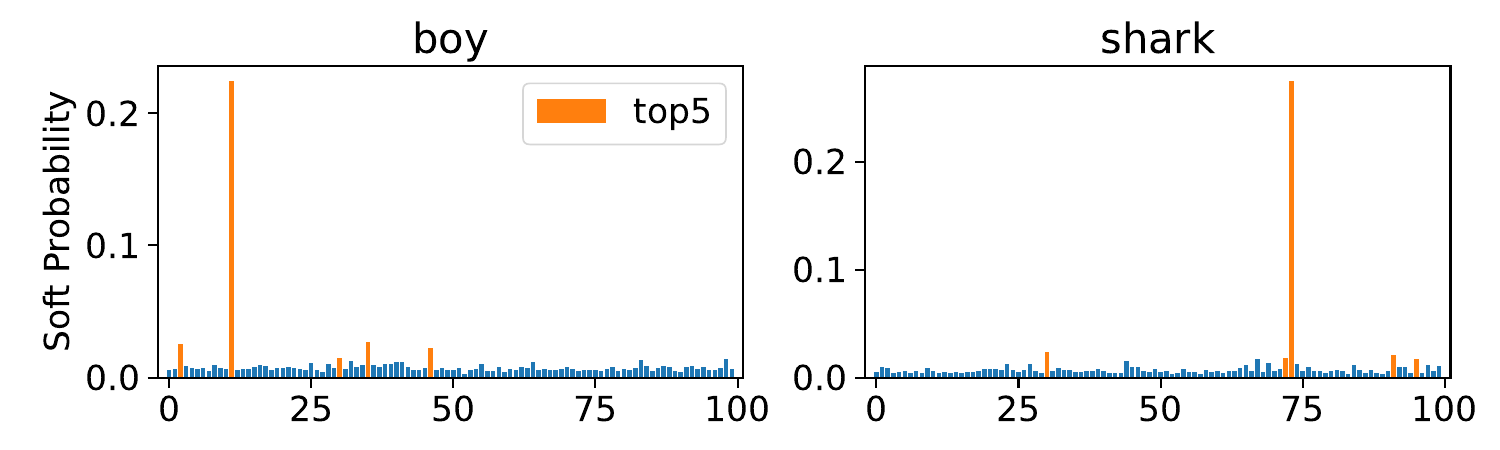}
    \label{fig:cifar100_probs_noaug}
  }\\
  \subfloat[]{
    \includegraphics[width=0.95\linewidth]{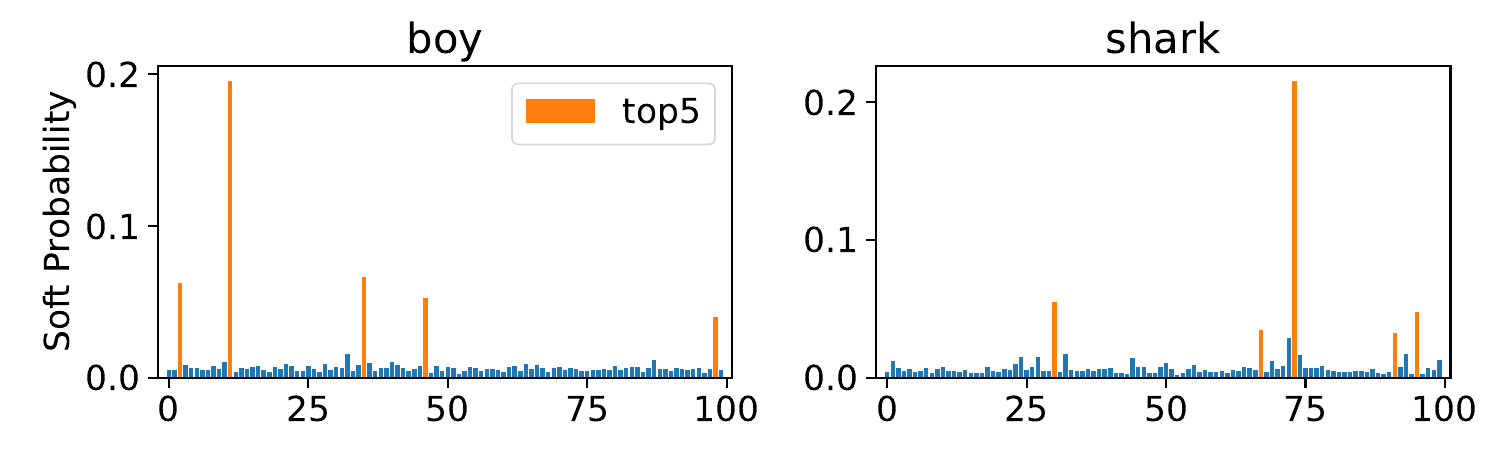}
    \label{fig:cifar100_probs_aug}
  }
  \caption{Average class predictions by teacher model ResNet32x4 on CIFAR-100. The figure compares the average predictions for two randomly selected classes, 'boy' and 'shark', using (a) standard data augmentation and (b) AutoAugment \cite{cubukAutoAugmentLearningAugmentation2019} on CIFAR-100's training set with a temperature of $T=4$. Top 5 predictions for each class are highlighted in different colors.}
  \label{fig:cifar100_probs}
\end{figure}

After the pre-training phase of the teacher model on CIFAR-100, its average soft predictions for each class are depicted in Fig.~\ref{fig:cifar100_probs_noaug}.
The results prominently display a unimodal distribution in the teacher model’s predictions, with the peak representing the highest average confidence for the most probable class.
Notably, the teacher model attained a 99.98\% accuracy on the CIFAR-100 training set.
This high accuracy suggests a strong correlation between the top logit and the target class, aligning with the original DKD's logit partition strategy.
The corresponding approximation of DKD within $\text{GDKD}_2$ selects the top logit of teacher predictions as the partition criterion. We denote it as GDKD-top1 and accordingly define the $\text{KL}(\bm{b}^\mathcal{T}||\bm{b}^\mathcal{S})$ component as \emph{Top Class Knowledge Distillation} (TopKD) and $\text{KL}(\bm{p}_{\setminus \mathbb{T}}^\mathcal{T}||\bm{p}_{\setminus \mathbb{T}}^\mathcal{S})$ as \emph{Other Classes Knowledge Distillation} (OtherKD) in Eq.~\eqref{eq:gdkd_loss_2}.

The gradients of the student logits $\bm{z}^\mathcal{S}=[z_1^\mathcal{S}, \dots, z_C^\mathcal{S}]$ with respect to $\mathcal{L}_{\text{GDKD-top1}}$ can be further decomposed into two components\footnote{For simplicity, the temperature $T$ is omitted (i.e., $T=1$) for concise formulation.}:
\begin{equation}
  \nabla_{\bm{z}^\mathcal{S}} \mathcal{L}_{\text{GDKD-top1}} = \nabla_{\bm{z}^\mathcal{S}} \mathcal{L}_\text{TopKD} + \beta \nabla_{\bm{z}^\mathcal{S}} \mathcal{L}_\text{OtherKD},
\end{equation}
\begin{equation}
  \begin{aligned}
    \nabla_{\bm{z}^\mathcal{S}} \mathcal{L}_\text{TopKD} & = \nabla_{\bm{z}^\mathcal{S}} \text{KL}(\bm{b}^\mathcal{T}||\bm{b}^\mathcal{S}) \\
                                                         & =
    \begin{cases}
      p_c^\mathcal{S}-p_c^\mathcal{T}                                                                         & i=c      \\
      p_i^\mathcal{S} \left[p_c^\mathcal{T}-\frac{\eta^\mathcal{T}}{\eta^\mathcal{S}} p_c^\mathcal{S} \right] & i \neq c
    \end{cases},
  \end{aligned}
  \label{eq:topkd_grad}
\end{equation}
\begin{equation}
  \begin{aligned}
    \nabla_{\bm{z}^\mathcal{S}} \mathcal{L}_\text{OtherKD} & = \nabla_{\bm{z}^\mathcal{S}} \text{KL}(\bm{p}_{\setminus \mathbb{T}}^\mathcal{T}||\bm{p}_{\setminus \mathbb{T}}^\mathcal{S}) \\
                                                           & =
    \begin{cases}
      0                                                                               & i=c      \\
      p_{\setminus \mathbb{T}, i}^\mathcal{S}-p_{\setminus \mathbb{T}, i}^\mathcal{T} & i \neq c
    \end{cases},
  \end{aligned}
  \label{eq:otherkd_grad}
\end{equation}
and $\eta=\sum_{i \neq c}p_i=1-p_c$ represents the sum of probabilities for non-top logits.

In comparison, the gradient of the student logits $\bm{z}^\mathcal{S}$ with respect to $\mathcal{L}_\text{KD}$ is:
\begin{equation}
  \begin{aligned}
    \nabla_{\bm{z}^\mathcal{S}} \mathcal{L}_\text{KD} & = \nabla_{\bm{z}^\mathcal{S}} \text{KL}(\bm{b}^\mathcal{T}||\bm{b}^\mathcal{S}) + (1-p_c^\mathcal{T}) \nabla_{\bm{z}^\mathcal{S}} \text{KL}(\bm{p}_{\setminus \mathbb{T}}^\mathcal{T}||\bm{p}_{\setminus \mathbb{T}}^\mathcal{S}) \\
                                                      & = p_i^\mathcal{S} -p_i^\mathcal{T}, \text{for }i=1,\dots,C.
  \end{aligned}
  \label{eq:kd_grad}
\end{equation}
This formulation indicates that GDKD-top1 retains the gradient of the top logit $z_c^\mathcal{S}$ as in the KD loss, while modifying the gradient weights of other logits. Specifically, the weight of $\nabla_{\bm{z}^\mathcal{S}} \mathcal{L}_\text{OtherKD}$ transitions from $1-p_c^\mathcal{T}$ to $\beta$.

To elucidate the specific contributions to gradients associated with non-top logits from the TopKD and OtherKD distillation terms, we utilize GDKD-top1 to train the student model. We then evaluate the average magnitudes of non-top logits' gradients for both terms, alongside their corresponding weights, as depicted in Fig.~\ref{fig:logits_grad_dist}. Throughout the training, the average magnitude of non-top logit gradients on $\mathcal{L}_\text{OtherKD}$ is significantly larger than that on $\mathcal{L}_\text{TopKD}$. Essentially, OtherKD dominates the model updates from non-top logits during distillation.
In contrast, the gradients of non-top logits in the original KD loss are suppressed due to the softmax operation, which exacerbates differences among logits, resulting in diminished prediction values for classes of non-top logits and impacting their gradients through Eq.~\eqref{eq:kd_grad}.

\begin{figure}
  \centering
  \subfloat[]{
    \includegraphics[width=0.95\linewidth]{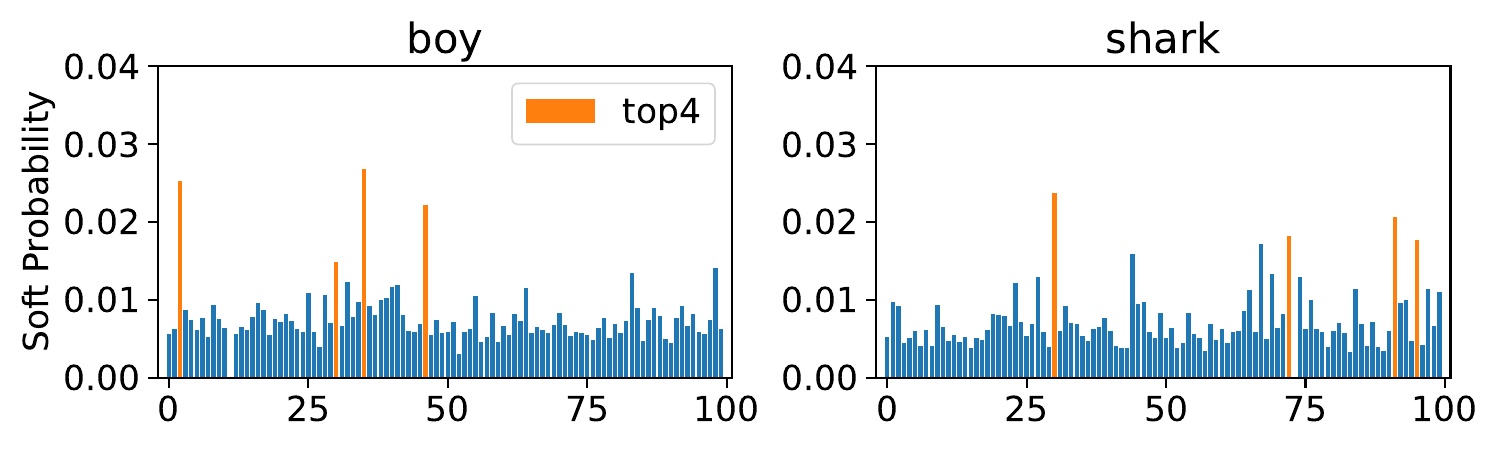}
  }\\
  \subfloat[]{
    \includegraphics[width=0.95\linewidth]{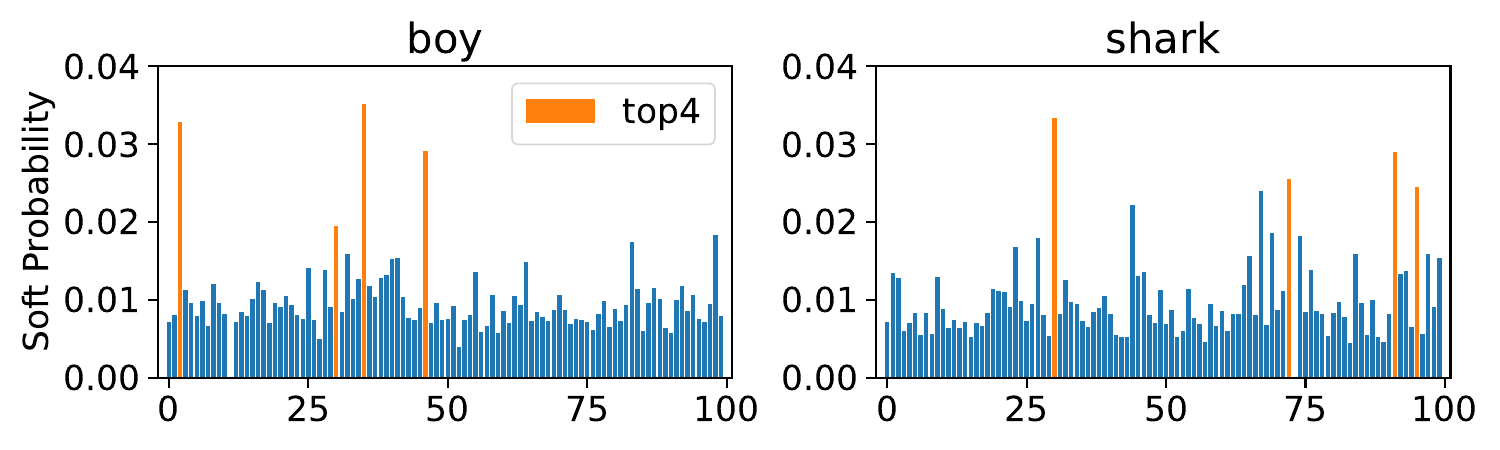}
  }
  \caption{The average predictions of teacher ResNet32x4 over one class of CIFAR-100's training dataset with temperature $T=4$, where the top1 predictions are ignored. (a): The predictions $\bm{p}^\mathcal{T}$ over one class without plotting the top1 value; (b): The reconstructed predictions of remaining logits $\bm{p}_{\setminus \mathbb{T}}^\mathcal{T}$ through softmax. The teacher is trained with standard data augmentations. The plot randomly selects 2 classes and marks the top 4 predictions with different colors.}
  \label{fig:cifar100_notop1_probs}
\end{figure}

In summary, the gradients of non-top logits in GDKD-top1 are largely governed by the OtherKD term, where its gradients $\beta(p_{\setminus \mathbb{T},i}^\mathcal{S}-p_{\setminus \mathbb{T},i}^\mathcal{T})$ are considerably larger than those in KD, i.e., $p_i^\mathcal{S} -p_i^\mathcal{T}$ for $i \in \setminus \mathbb{T}$.
Since these non-top logits, encompassing a majority of classes, are typically viewed as repositories of valuable \emph{dark knowledge} \cite{hintonDistillingKnowledgeNeural2015}, increasing their gradients' magnitude plays a crucial role in augmenting the student model's ability to extract information from non-top logits of the teacher model. The increase of the gradients' magnitude comes from two factors.
On the one hand, the teacher model's recalculated softmax predictions on these non-top logits, denoted as $\bm{p}_{\setminus \mathbb{T}}^\mathcal{T}$, exceed the original softmax predictions $\bm{p}^\mathcal{T}$ used in KD:
\begin{equation}
  \begin{aligned}
    \forall i \in \setminus \mathbb{T},\quad p_{\setminus \mathbb{T},i}^\mathcal{T} & > p_{i}^\mathcal{T}                                                                                                \\
                                                                                    & = \frac{\exp(z_i^\mathcal{T})}{\exp(z_c^\mathcal{T}) + \sum_{i \in {\setminus \mathbb{T}}} \exp(z_i^\mathcal{T})},
  \end{aligned}
\end{equation}
since $\exp(z_c^\mathcal{T}) > 0$. In $\text{GDKD}_2$, when $\mathbb{T}=\{c\}$ includes the class of the teacher's top logit, it holds that $\exp(z_c) \ge \exp(z_i)$ for $i \in \{1,\dots,C\}$. And when $c$ is the class with the highest probability from the teacher, the maximum enhancement of $p_{\setminus \mathbb{T},i}^\mathcal{T}$ for $i \in \setminus \mathbb{T}$ achieves, indirectly improving the gradient magnitudes of non-top logits. Fig.~\ref{fig:cifar100_notop1_probs} demonstrates how the recalculated probabilities $p_{\setminus \mathbb{T},i}^\mathcal{T}$ increase from $p_{i}^\mathcal{T}$ when the top logit class is chosen as the partition strategy.
On the other hand, the large coefficient $\beta$ attached to OtherKD amplifies the gradients of non-top logits, thereby further enhancing the knowledge extraction from these logits.


In conclusion, the effectiveness of the $\text{GDKD}_2$ approach, especially when using the partition strategy $c = \arg\max_i (z_i^\mathcal{T})$, hinges on two crucial aspects in logit-based decoupled knowledge distillation methods:
\begin{itemize}
  \item \textbf{Augmented logits interrelationship via top-logit partitioning}: A key finding is that partitioning logits according to the top logit significantly improves the expression of knowledge from non-top logits. By isolating the top logit, the gradients of non-top logits are no longer suppressed by the top logit through the softmax operation, thus allowing for more effective knowledge transfer via the enhanced relations among them. This suppression is often overlooked in previous logit-based knowledge distillation methods.
  \item \textbf{Enhanced knowledge extraction through adjusted large weight}: Another critical insight is that increasing the weight beyond the coupled one on the non-top loss component, can further amplify the knowledge extraction among non-top logits during distillation. This adjustment in weighting empowers the student model to assimilate and utilize the intricate relationships and insights inherent in these logits that share small probabilities, leading to a more enriched and comprehensive learning experience.
\end{itemize}

\subsection{GDKD Algorithm}
\label{sec:algorithm}

\begin{algorithm}[t]
  \caption{Pseudocode of GDKD}
  \label{algo:gdkd_topk_impl}
  \begin{algorithmic}
    \State \textbf{Input:} student logits $\bm{z}^\mathcal{S}$, teacher logits $\bm{z}^\mathcal{T}$, hyperparameters of GDKD: $k, w_0, w_1, w_2$, temperature $T$
    \LineComment{get the top-$k$ classes}
    \State $i_\text{topk} = \text{topk}(\bm{z}^\mathcal{T}, k).\text{indices}$
    \LineComment{get the other classes}
    \State $i_\text{other} = \text{setminus} (i_\text{topk})$
    \State
    \State $\bm{p}^\mathcal{T} = \text{softmax}(\bm{z}^\mathcal{T}/T)$
    \State $\bm{p}^\mathcal{S} = \text{softmax}(\bm{z}^\mathcal{S}/T)$
    \State $\bm{b}^\mathcal{T} = \left[\text{sum} (p^\mathcal{T}[i_\text{topk}]), \text{sum} (p^\mathcal{T}[i_\text{other}]) \right]$
    \State $\bm{b}^\mathcal{S} = \left[\text{sum} (p^\mathcal{S}[i_\text{topk}]), \text{sum} (p^\mathcal{S}[i_\text{other}]) \right]$
    \State $\bm{p}_\text{topk}^\mathcal{T} = \text{softmax}(\bm{z}^\mathcal{T}[i_\text{topk}]/T)$
    \State $\bm{p}_\text{topk}^\mathcal{S} = \text{softmax}(\bm{z}^\mathcal{S}[i_\text{topk}]/T)$
    \State $\bm{p}_\text{other}^\mathcal{T} = \text{softmax}(\bm{z}^\mathcal{T}[i_\text{other}]/T)$
    \State $\bm{p}_\text{other}^\mathcal{S} = \text{softmax}(\bm{z}^\mathcal{S}[i_\text{other}]/T)$
    \State
    \LineComment{calculate decoupled KD loss terms}
    \State $\mathcal{L}_\text{high} = \text{KL}(\bm{b}^\mathcal{T}||\bm{b}^\mathcal{S})$
    \State $\mathcal{L}_\text{low-topk} = \text{KL}(\bm{p}_\text{topk}^\mathcal{T}||\bm{p}_\text{topk}^\mathcal{S})$
    \State $\mathcal{L}_\text{low-other} = \text{KL}(\bm{p}_\text{other}^\mathcal{T}||\bm{p}_\text{other}^\mathcal{S})$
    \State $\mathcal{L}_\text{GDKD} = w_0 \mathcal{L}_\text{high} + w_1 \mathcal{L}_\text{low-topk} + w_2 \mathcal{L}_\text{low-other}$
  \end{algorithmic}
\end{algorithm}

\begin{figure}[t]
  \centering

  \subfloat{
    \includegraphics[width=0.9\linewidth]{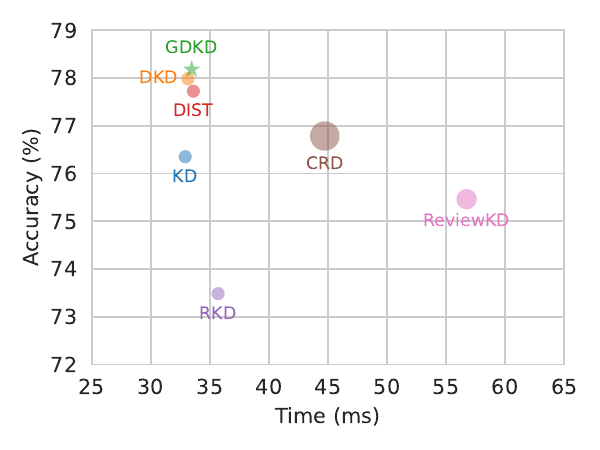}
  }\\
  \subfloat{
    \includegraphics[width=0.9\linewidth]{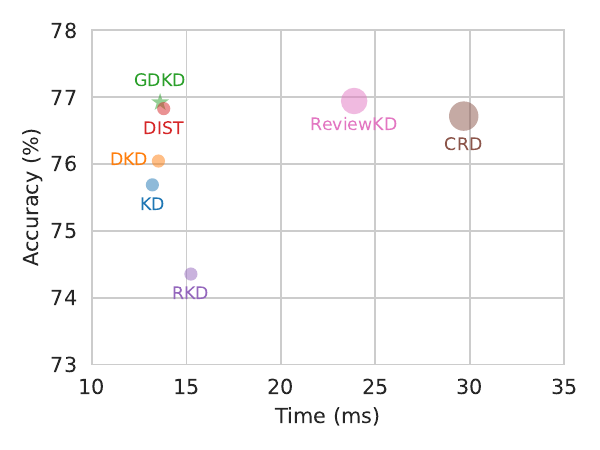}
  }
  \caption{Comparison of average iteration training speed and accuracy for various knowledge distillation methods on CIFAR-100, with two distinct tasks presented. \textbf{Top:} $\mathcal{T}$: ResNet32x4, $\mathcal{S}$: ShuffleNet-V1; \textbf{Bottom:} $\mathcal{T}$: WRN-40-2 and $\mathcal{S}$: WRN-16-2. The size of each point in the graph correlates with the number of additional parameters required by each distillation method. Notably, KD, DKD, DIST, and GDKD ($k=5$) do not necessitate extra parameters. However, the following methods require them. \textbf{Top}: CRD requires 12.96M and ReviewKD requires 4.61M; \textbf{Bottom}: CRD needs 12.83M and ReviewKD needs 0.45M.}
  \label{fig:train_speed}
\end{figure}

Leveraging the foundational insights as investigated above, we present the \emph{GDKD algorithm} with an innovative partitioning strategy.
The motivation stems from our observations of the teacher models' predictive distributions: under different datasets or training settings, the teacher models may exhibit multimodal distributions in predictions, as depicted in Fig.~\ref{fig:cifar100_probs}. In this pattern, the prediction distribution of a sample presents multiple high-probability predicted classes, which is often deemed as an embodiment of the teacher model's ability to distill the intrinsic linkages among classes in the dataset.
However, in multimodal scenarios, using the top logit partitioning strategy could inadvertently partition \emph{small logits}\footnote{For clarity, we use \emph{small logits} referred to as logits associated with distinctly small probabilities of the teacher under a temperature $T$.} together with other non-top logits with high probabilities in the same group. In consequence, relationships among small logits might be diluted as these large non-top logits might decrease the probability of small logits through softmax.

Therefore, our primary proposition is to partition logits into clusters with comparable sizes under associated soft prediction probabilities.
Specifically, for every input instance, we divide the teacher and student logits based on values of the teacher's predictive distribution: one group for the majority of logits with lower probabilities and the other for the higher ones. Then the bifurcation is leveraged to calculate the GDKD loss in Eq.~\eqref{eq:gdkd_loss_3}. This process ensures that relationships derived from the recalculated softmax predictions of small logits are optimally enhanced.


Pseudocode of the algorithm is detailed in Algorithm~\ref{algo:gdkd_topk_impl}.
We implement the above partition strategy in a straightforward way, choosing a predetermined number of top-$k$ classes from teacher logits to comprise $\mathbb{T}$, and the remaining classes are then allocated to $\setminus \mathbb{T}$.
This method keeps the efficiency in KD \cite{hintonDistillingKnowledgeNeural2015} and DKD \cite{zhaoDecoupledKnowledgeDistillation2022}, relying solely on the search for unordered top-$k$ classes and achieving a computational complexity of $O(n)$ when $n \gg k$.
As depicted in Fig.~\ref{fig:train_speed}, despite its simplicity, the proposed GDKD algorithm adeptly strikes a balance between model performance and training speed.
Additionally, it avoids the need for extra parameters, thereby setting itself apart from these sophisticated feature-based methods.

\section{Experiments}
\label{sec:exp}

In this section, we comprehensively assess the performance of GDKD on a variety of tasks, including image classification tasks, transfer learning tasks, and semantic segmentation tasks.
We also demonstrate our ablation studies on the hyperparameters and loss terms.
Furthermore, we conduct additional extended experiments to further explore the efficacy of GDKD.




\subsection{Experimental Settings}

\subsubsection{Image Classification Experiments}

The image classification experiments are conducted using the following two datasets.

\textbf{CIFAR-100} \cite{krizhevskyLearningMultipleLayers} comprises 32x32 pixel images spanning 100 classes for classification. The dataset consists of 50k training images and 10k test images. Due to the limited size of the training set, overfitting often emerges as a prevalent concern, which could hinder the effect of knowledge from teacher models. To counteract this, we employ AutoAugment \cite{cubukAutoAugmentLearningAugmentation2019} as a robust data augmentation technique during the training phases for teacher models.
The training of student models remains the standard data augmentation procedure: random crop + horizontal flip. Notably, we discern that, with AutoAugment, pre-trained teachers would output predictions in multimodal distributions, as depicted in Fig.~\ref{fig:cifar100_probs_aug}. Consequently, this dataset serves as an apt benchmark for assessing the efficacy of GDKD, where an optimal value of $k > 1$ is favored. For logit-based knowledge distillation methods, we set the temperature $T$ to 4 during the experiments.

\begin{figure*}[t]
  \centering
  \subfloat{
    \includegraphics[width=0.95\linewidth]{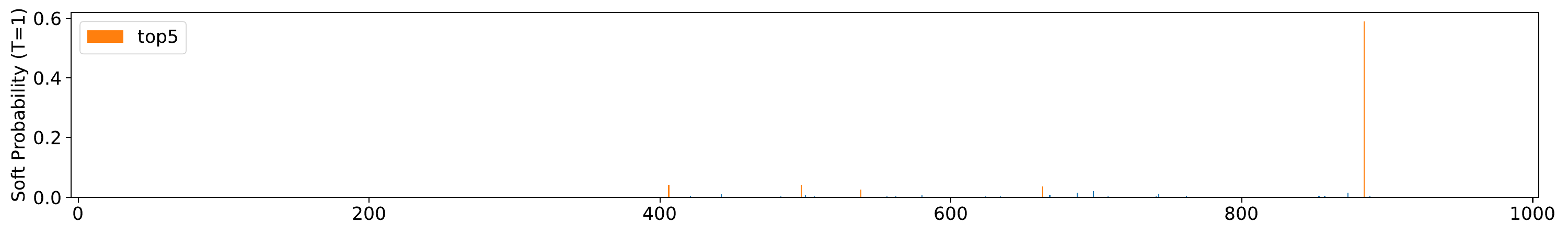}
  }\\
  \subfloat{
    \includegraphics[width=0.95\linewidth]{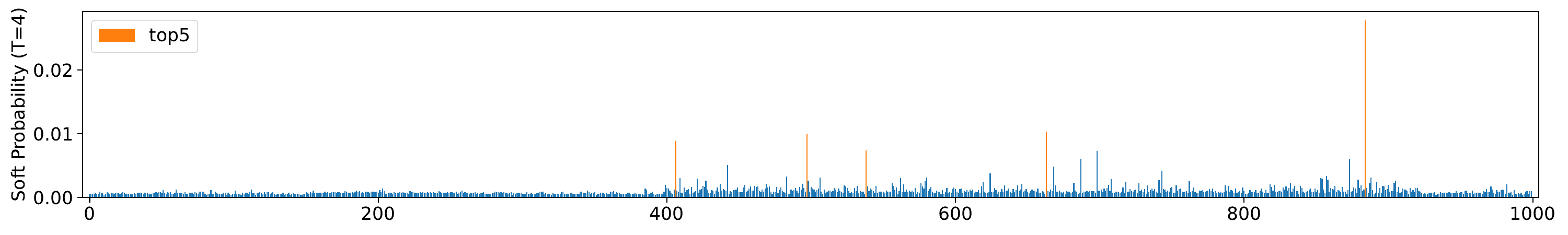}
  }
  \caption{The average predictions of the teacher model (ResNet34) over one class of ImageNet's training set. \textbf{Top:} The temperature is set to $T=1$, which is used in our experiments.  \textbf{Bottom:} The average predictions are softened by temperature $T=4$ for better observation of probability discrepancy.}
  \label{fig:imagenet_probs}
\end{figure*}

\textbf{ImageNet} \cite{russakovskyImageNetLargeScale2015} is a vast image classification dataset, encompassing 1.2 million training images and 50k validation images across 1000 classes. We note that when subjected to standard data augmentation procedures, the teacher's predictions also appear multimodal, as illustrated in Fig.~\ref{fig:imagenet_probs}.

\subsubsection{Transfer Learning Experiments}
Transfer learning experiments are conducted using the following two datasets.
In these experiments, the teacher and student models are pre-trained from scratch on the ImageNet dataset. Then the teacher model is pre-fine-tuned on the target dataset, which is later utilized to fine-tune the student model by different distillation methods.

\begin{table*}[b]
  \centering
  \caption{Results of Distillation Methods on the CIFAR-100 Test Set for Pairs of Teacher and Student with Homogeneous Architectures}
  \label{table:cifar100_aug_results_same_arch}
  \begin{threeparttable}
    \begin{tabular}{*{9}{c}}
      \toprule
      \multirow{4}{*}{Type}     & \multirow{4}{*}{Method}                                  & \multirow{2}{*}{$\mathcal{T}$} & ResNet56       & ResNet110      & ResNet32x4     & WRN-40-2       & WRN-40-2       & VGG13          \\
                                &                                                          &                                & 74.26          & 76.55          & 82.05          & 77.95          & 77.95          & 76.08          \\
                                &                                                          & \multirow{2}{*}{$\mathcal{S}$} & ResNet20       & ResNet32       & ResNet8x4      & WRN-16-2       & WRN-40-1       & VGG8           \\
                                &                                                          &                                & 69.06          & 71.14          & 72.50          & 73.26          & 71.98          & 70.36          \\ \midrule
      \multirow{3}{*}{features} & RKD \cite{parkRelationalKnowledgeDistillation2019}       &                                & 70.81          & 73.09          & 72.47          & 74.35          & 73.13          & 71.94          \\
                                & CRD \cite{tianContrastiveRepresentationDistillation2022} &                                & 71.93          & 74.89          & 76.63          & 76.72          & 75.66          & 74.92          \\
                                & ReviewKD \cite{chenDistillingKnowledgeKnowledge2021}     &                                & \textbf{72.00} & \textbf{75.46} & 75.33          & \textbf{76.94} & \textbf{76.00} & 72.79          \\ \midrule
      \multirow{6}{*}{logits}   & KD \cite{hintonDistillingKnowledgeNeural2015}            &                                & 70.16          & 73.58          & 74.68          & 75.69          & 74.62          & 74.44          \\
                                & DKD \cite{zhaoDecoupledKnowledgeDistillation2022}        &                                & 71.15          & 74.35          & 76.64          & 76.04          & 75.20          & 75.30          \\
                                & DIST \cite{huangKnowledgeDistillationStronger2022}       &                                & 71.27          & 74.47          & 77.59          & 76.83          & 75.58          & 75.28          \\
                                & KD w/ LS \cite{sunLogitStandardizationKnowledge2024}     &                                & 70.14          & 73.81          & \textbf{78.36} & 76.24          & 75.11          & 75.24          \\
                                & \textbf{GDKD}                                            &                                & 71.65          & 74.82          & 77.76          & \textbf{76.92} & 75.75          & \textbf{75.74} \\
                                & $\Delta$                                                 &                                & +0.50          & +0.47          & +1.12          & +0.88          & +0.55          & +0.44          \\ \bottomrule
    \end{tabular}
    \begin{tablenotes}
      \item $\Delta$ represents the performance improvement of GDKD from DKD. All results are averaged over 5 independent runs.
    \end{tablenotes}
  \end{threeparttable}
\end{table*}

\textbf{Tiny-Imagenet} \cite{Le2015TinyIV} is a subset of the ImageNet dataset downsampling to 64x64. The dataset includes 200 classes, where each class has 500 images for training and 50 images for testing. We use the same standard data augmentation for both teachers and students during the fine-tuning.

\textbf{CUB-200-2011} \cite{WahCUB_200_2011} is a fine-grained dataset, comprising 200 classes of birds, with 5994 images for training and 5794 images for testing. And since the size of the training set is limited, we also employ a strong data augmentation strategy at the teacher fine-tuning stage.

\subsubsection{Semantic Segmentation Experiments}
The semantic segmentation experiments are conducted using the following dataset.

\textbf{Cityscapes} \cite{Cordts2016Cityscapes} is a large-scale dataset for image semantic segmentation tasks, containing 2975 training images and 500 validation images. Each image has a pixel-level annotated label image, where the number of classes in the task is 19.
Additionally, we find that the relations among the local area pixels are essential in semantic segmentation tasks, therefore we append the intra-class relation loss in DIST \cite{huangKnowledgeDistillationStronger2022} for all logit-based methods, ensuring fair comparisons.

\subsubsection{Training Settings}
As discussed in Section~\ref{sec:algorithm}, $k$ is a hyperparameter in the proposed GDKD algorithm. The ablation study in Section~\ref{sec:ablation_k} indicates that GDKD's performance is relatively unaffected by the choice of $k$ within a rational range. Consequently, we consistently set $k$ to 5 for all tasks, as this value satisfactorily covers our experimental requirements and yields optimal performance, based on our observations. Detailed implementation specifics are further elaborated in Appendix~\ref{sec:train_impl}.

\subsection{Main Results}

\begin{table*}[t]
  \centering
  \caption{Results of Distillation Methods on the CIFAR-100 Test Set for Pairs of Teacher and Student with Heterogeneous Architectures}
  \label{table:cifar100_aug_results_diff_arch}
  \begin{threeparttable}
    \begin{tabular}{*{8}{c}}
      \toprule
      \multirow{4}{*}{Type}     & \multirow{4}{*}{Method}                                  & \multirow{2}{*}{$\mathcal{T}$} & ResNet32x4     & ResNet32x4     & WRN-40-2       & ResNet50       & VGG13          \\
                                &                                                          &                                & 82.05          & 82.05          & 77.95          & 80.98          & 76.08          \\
                                &                                                          & \multirow{2}{*}{$\mathcal{S}$} & ShuffleNet-V1  & ShuffleNet-V2  & ShuffleNet-V1  & MobileNet-V2   & MobileNet-V2   \\
                                &                                                          &                                & 70.50          & 71.82          & 70.50          & 64.60          & 64.60          \\ \midrule
      \multirow{3}{*}{features} & RKD \cite{parkRelationalKnowledgeDistillation2019}       &                                & 73.49          & 74.62          & 74.39          & 65.68          & 65.93          \\
                                & CRD \cite{tianContrastiveRepresentationDistillation2022} &                                & 76.79          & 77.86          & 77.52          & 69.59          & \textbf{70.43} \\
                                & ReviewKD \cite{chenDistillingKnowledgeKnowledge2021}     &                                & 75.46          & 76.58          & 77.55          & 65.72          & 67.17          \\ \midrule
      \multirow{6}{*}{logits}   & KD \cite{hintonDistillingKnowledgeNeural2015}            &                                & 76.35          & 77.25          & 77.06          & 68.82          & 68.36          \\
                                & DKD \cite{zhaoDecoupledKnowledgeDistillation2022}        &                                & 77.99          & 78.62          & 77.49          & 69.88          & 70.03          \\
                                & DIST \cite{huangKnowledgeDistillationStronger2022}       &                                & 77.72          & 78.86          & 77.89          & 70.01          & 69.55          \\
                                & KD w/ LS \cite{sunLogitStandardizationKnowledge2024}     &                                & 77.64          & 78.34          & 77.55          & 70.14          & 69.82          \\
                                & \textbf{GDKD}                                            &                                & \textbf{78.21} & \textbf{79.27} & \textbf{78.05} & \textbf{70.55} & 70.35          \\
                                & $\Delta$                                                 &                                & +0.22          & +0.65          & +0.56          & +0.67          & +0.32          \\ \bottomrule
    \end{tabular}
    \begin{tablenotes}
      \item $\Delta$ represents the performance improvement of GDKD from DKD. All results are averaged over 5 independent runs.
    \end{tablenotes}
  \end{threeparttable}
\end{table*}

\begin{table*}[t]
  \centering
  \caption{Results of Different Distillation Methods on the ImageNet Validation Set}
  \label{table:imagenet_results}
  \begin{threeparttable}
    \begin{tabular}{*{7}{c}}
      \toprule
      Type                      & Method                                                            & Top-1 Accuracy & Top-5 Accuracy & Method                                                            & Top-1 Accuracy & Top-5 Accuracy \\ \midrule
                                & $\mathcal{T}$: ResNet34                                           & 73.31          & 91.42          & $\mathcal{T}$: ResNet50                                           & 76.16          & 92.87          \\
                                & $\mathcal{S}$: ResNet18                                           & 69.75          & 89.07          & $\mathcal{S}$: MobileNet-V1                                       & 68.87          & 88.76          \\ \midrule
      \multirow{4}{*}{features} & AT\tnote{*} \cite{zagoruykoPayingMoreAttention2022}               & 70.69          & 90.01          & AT\tnote{*} \cite{zagoruykoPayingMoreAttention2022}               & 69.56          & 89.33          \\
                                & OFD\tnote{*} \cite{heoComprehensiveOverhaulFeature2019}           & 70.81          & 89.98          & OFD\tnote{*} \cite{heoComprehensiveOverhaulFeature2019}           & 71.25          & 90.34          \\
                                & CRD\tnote{*} \cite{tianContrastiveRepresentationDistillation2022} & 71.17          & 90.13          & CRD\tnote{*} \cite{tianContrastiveRepresentationDistillation2022} & 71.37          & 90.41          \\
                                & ReviewKD\tnote{*} \cite{chenDistillingKnowledgeKnowledge2021}     & 71.61          & 90.51          & ReviewKD\tnote{*} \cite{chenDistillingKnowledgeKnowledge2021}     & 72.56          & 91.00          \\ \midrule
      \multirow{6}{*}{logits}   & KD\tnote{*} \cite{hintonDistillingKnowledgeNeural2015}            & 71.03          & 90.05          & KD\tnote{*} \cite{hintonDistillingKnowledgeNeural2015}            & 70.50          & 89.80          \\
                                & DKD\tnote{*} \cite{zhaoDecoupledKnowledgeDistillation2022}        & 71.70          & 90.41          & DKD\tnote{*} \cite{zhaoDecoupledKnowledgeDistillation2022}        & 72.05          & 91.05          \\
                                & DIST \cite{huangKnowledgeDistillationStronger2022}                & 71.57          & 90.68          & DIST \cite{huangKnowledgeDistillationStronger2022}                & 73.09          & \textbf{91.40} \\
                                & KD w/ LS\tnote{*} \cite{sunLogitStandardizationKnowledge2024}     & 71.42          & 90.29          & KD w/ LS\tnote{*} \cite{sunLogitStandardizationKnowledge2024}     & 72.18          & 90.80          \\
                                & \textbf{GDKD}                                                     & \textbf{72.26} & \textbf{90.72} & \textbf{GDKD}                                                     & \textbf{73.25} & 91.35          \\
                                & $\Delta$                                                          & +0.56          & +0.31          & $\Delta$                                                          & +1.20          & +0.30          \\\bottomrule
    \end{tabular}
    \begin{tablenotes}
      \item [*] The results are from DKD \cite{zhaoDecoupledKnowledgeDistillation2022} and LS \cite{sunLogitStandardizationKnowledge2024}.
      \item $\Delta$ represents the performance improvement of GDKD from DKD. All results are the average of 3 independent runs.
    \end{tablenotes}
  \end{threeparttable}
\end{table*}

\begin{table*}[t]
  \centering
  \caption{Transfer Learning Results of Different Distillation Methods on the Tiny-ImageNet and CUB-200-2011 Test Set}
  \label{table:transfer_learning_results}
  \begin{threeparttable}
    \begin{tabular}{*{7}{c}}

      \multicolumn{3}{c}{Dataset} & \multicolumn{2}{c}{Tiny-ImageNet}                    & \multicolumn{2}{c}{CUB-200-2011}                                                                                                                                \\ \toprule
      \multirow{4}{*}{Type}       & \multirow{4}{*}{Method}                              & \multirow{2}{*}{$\mathcal{T}$}   & ResNet34                        & ResNet50                        & ResNet34                        & ResNet50               \\
                                  &                                                      &                                  & 77.04 (92.62)                   & 78.23 (93.03)                   & 80.45 (95.93)                   & 83.00 (96.74)          \\
                                  &                                                      & \multirow{2}{*}{$\mathcal{S}$}   & ResNet18                        & MobileNet-V1                    & ResNet18                        & MobileNet-V1           \\
                                  &                                                      &                                  & 73.28 (91.10)                   & 71.14 (90.07)                   & 78.07 (94.63)                   & 77.81 (94.77)          \\ \midrule
      features                    & ReviewKD \cite{chenDistillingKnowledgeKnowledge2021} &                                  & 74.12 (91.61)                   & 71.95 (90.38)                   & 80.33 (95.67)                   & 80.56 (95.87)          \\ \midrule
      \multirow{4}{*}{logits}     & KD \cite{hintonDistillingKnowledgeNeural2015}        &                                  & 74.13 (91.58)                   & 71.45 (90.33)                   & 79.09 (95.23)                   & 79.03 (95.20)          \\
                                  & DKD \cite{zhaoDecoupledKnowledgeDistillation2022}    &                                  & 74.51 (91.52)                   & 72.19 (90.34)                   & 80.25 (95.65)                   & 80.53 (\text{96.17})   \\
                                  & DIST \cite{huangKnowledgeDistillationStronger2022}   &                                  & 73.94 (91.09)                   & 71.98 (90.27)                   & 80.38 (95.59)                   & 80.49 (95.62)          \\
                                  & \textbf{GDKD}                                        &                                  & \textbf{74.79} (\textbf{91.73}) & \textbf{72.75} (\textbf{90.58}) & \textbf{81.11} (\textbf{95.84}) & \textbf{81.82} (95.87) \\ \bottomrule
    \end{tabular}
    \begin{tablenotes}
      \item The student and teacher models are pre-trained on the ImageNet. The Top-1 and Top-5 accuracy are collected from the average of 4 independent runs.
    \end{tablenotes}
  \end{threeparttable}
\end{table*}

\subsubsection{Results on CIFAR-100}
In this part, we detail the performance of GDKD and compare it with other KD methods across a range of distillation tasks on CIFAR-100.
Our results are presented in two distinct contexts: where teacher and student models share the same architecture (Table~\ref{table:cifar100_aug_results_same_arch}) and where they have different architectures (Table~\ref{table:cifar100_aug_results_diff_arch}).

A key observation is when teachers generate multimodal predictions, GDKD's consistent outperformance over DKD in all tasks.
This result supports our analysis of the logit decoupling mechanism, demonstrating GDKD's superior ability to efficiently utilize knowledge from small logits in multimodal scenarios.

Furthermore, GDKD's performance surpasses more than half of the 11 evaluated tasks against both the latest logit-based and feature-based knowledge distillation methods and matches the performance of other methods in the rest of the tasks.
This achievement highlights the potential of logit-based distillation in advancing the field.
An interesting aspect of GDKD’s performance is its enhanced effectiveness in tasks with heterogeneous architectures.
This advantage may arise from the limitations of feature-based rivals, which are often sensitive to the architectural congruence between teacher and student models.
Unlike feature-based methods that rely on architecture-specific knowledge and require adapter modules for dimension matching, GDKD capitalizes on the final output predictions, making it more adaptable to varying architectures.

\subsubsection{Results on ImageNet}
The performance of various knowledge distillation methods on the ImageNet dataset is comprehensively summarized in Table~\ref{table:imagenet_results}.
In this comparison, GDKD notably distinguishes itself by outperforming all other evaluated methods.
This result reinforces the strength of GDKD, particularly in handling multimodal prediction scenarios, and further emphasizes the growing significance and versatility of logit-based knowledge in advanced machine learning applications.

\subsubsection{Results on Tiny-ImageNet \& CUB-200-2011}
\label{sec:transfer_learning_results}
Our evaluation of Tiny-ImageNet and CUB-200-2011, known for the high similarity between classes, showcases GDKD's capability in Table~\ref{table:transfer_learning_results}.
GDKD excels in extracting and transferring nuanced relationships between classes, especially for the classes with small logits, consistently outperforming other distillation methods, including advanced feature-based techniques.
A highlight is GDKD's performance on CUB-200-2011, where a GDKD-distilled student model (ResNet18) surpasses the performance of its teacher model (ResNet34).
This result underlines GDKD's effectiveness and its potential in logit-based knowledge extraction and transfer.

\subsection{Results on Cityscapes}
In the case of semantic segmentation with Cityscapes, we use DeepLabV3 \cite{chenEncoderDecoderAtrousSeparable2018} with a ResNet101 backbone as the teacher model and compare it with student models based on either DeepLabV3 or PSPNet \cite{zhaoPyramidSceneParsing2017}, both with a ResNet18 backbone.
As detailed in Table~\ref{table:cityscapes_results}, GDKD consistently outperforms other distillation methods in this challenging domain, including methods specific to semantic segmentation tasks.
The performance not only underscores the versatility of GDKD but also suggests that logit-based knowledge distillation, particularly in our GDKD formulation, can be more effective than feature-based methods in complex tasks such as semantic segmentation.

\begin{table}[t]
  \centering
  \caption{mIoU (\%) of Different Distillation Methods on the Cityscapes Validation Set}
  \label{table:cityscapes_results}
  \resizebox{\linewidth}{!}{
    \begin{threeparttable}
      \begin{tabular}{*{7}{c}}
        \toprule
        \multirow{4}{*}{Type}     & \multirow{4}{*}{Method}                                     & DeepLabV3-R101 & DeepLabV3-R101 \\
                                  &                                                             & 78.07          & 78.07          \\
                                  &                                                             & DeepLabV3-R18  & PSPNet-R18     \\
                                  &                                                             & 74.21          & 72.55          \\ \midrule
        \multirow{4}{*}{features} & SKD\tnote{*} \cite{liuStructuredKnowledgeDistillation2023}  & 75.42          & 73.29          \\
                                  & IFVD\tnote{*} \cite{wangIntraclassFeatureVariation2020}     & 75.59          & 73.71          \\
                                  & CWD\tnote{*} \cite{shuChannelWiseKnowledgeDistillation2021} & 75.55          & 74.36          \\
                                  & CIRKD\tnote{*} \cite{yangCrossImageRelationalKnowledge2022} & 76.38          & 74.73          \\ \midrule
        \multirow{3}{*}{logits}   & KD\tnote{+} \cite{hintonDistillingKnowledgeNeural2015}      & 77.03          & 75.46          \\
                                  & DKD\tnote{+} \cite{zhaoDecoupledKnowledgeDistillation2022}  & 76.99          & 74.67          \\
                                  & DIST \cite{huangKnowledgeDistillationStronger2022}          & 76.76          & 75.00          \\
                                  & \textbf{GDKD}\tnote{+}                                      & \textbf{77.42} & \textbf{75.62} \\ \bottomrule
      \end{tabular}
      \begin{tablenotes}
        \item [*] The results are from CIRKD \cite{yangCrossImageRelationalKnowledge2022} and DIST \cite{zhaoDecoupledKnowledgeDistillation2022}.
        \item [+] The logit-based method is incorporated with the intra-class relation loss from DIST \cite{huangKnowledgeDistillationStronger2022}.
        \item All results are the average of 3 independent runs.
      \end{tablenotes}
    \end{threeparttable}
  }
\end{table}

\subsection{Ablation Studies}

\begin{table*}[t]
  \centering
  \caption{Ablation Study Results of Different Components in GDKD on the CIFAR-100 Test Set.}
  \label{table:gdkd_terms_ablation}
  \begin{threeparttable}
    \begin{tabular}{ccc|llll}
      \toprule
      \multirow{2}{*}{highKD}                                            & \multirow{2}{*}{lowKD-topk} & \multirow{2}{*}{lowKD-other} & \multicolumn{2}{c}{$\mathcal{T}$: ResNet32x4 82.05} & \multicolumn{2}{c}{$\mathcal{T}$: WRN-40-2 77.95}                                                          \\
                                                                         &                             &                              & $\mathcal{S}$: ResNet8x4                            & $\mathcal{S}$: ShuffleNet-V2                      & $\mathcal{S}$: WRN-40-1 & $\mathcal{S}$: ShuffleNet-V1 \\ \midrule
                                                                         &                             &                              & 72.50                                               & 71.82                                             & 71.98                   & 70.50                        \\
      \checkmark                                                         &                             &                              & 70.56 (-1.94)                                       & 74.16 (+2.34)                                     & 72.02 (+0.04)           & 73.64 (+3.14)                \\
                                                                         & \checkmark                  &                              & 72.42 (-0.08)                                       & 73.12 (+1.30)                                     & 72.18 (+0.20)           & 74.42 (+3.92)                \\
                                                                         &                             & \checkmark                   & 76.03 (+3.53)                                       & 79.22 (+7.40)                                     & 75.32 (+3.34)           & 77.99 (+7.49)                \\
      \checkmark                                                         & \checkmark                  &                              & 73.35 (+0.85)                                       & 75.58 (+3.76)                                     & 74.08 (+2.10)           & 76.33 (+5.83)                \\
      \checkmark                                                         &                             & \checkmark                   & 76.16 (+3.66)                                       & 78.77 (+6.95)                                     & 75.02 (+3.04)           & 77.91 (+7.41)                \\
                                                                         & \checkmark                  & \checkmark                   & 77.71 (+5.21)                                       & 79.12 (+7.30)                                     & 75.55 (+3.57)           & 77.99 (+7.49)                \\
      \checkmark                                                         & \checkmark                  & \checkmark                   & \textbf{77.76 (+5.26)}                              & \textbf{79.27 (+7.45)}                            & \textbf{75.75 (+3.77)}  & \textbf{78.05 (+7.55)}       \\ \midrule
      \multicolumn{3}{c|}{KD \cite{hintonDistillingKnowledgeNeural2015}} & 74.68 (+2.18)               & 76.35 (+5.85)                & 74.62 (+2.64)                                       & 77.25 (+5.43)                                                                                              \\ \bottomrule
    \end{tabular}
    \begin{tablenotes}
      \item The accuracy of KD is also reported for comparison, and all results are the mean of 5 runs. Additionally, the improvement from the student model trained from scratch is reported.
    \end{tablenotes}
  \end{threeparttable}
\end{table*}

\begin{table*}[t]
  \centering
  \caption{Results of GDKD and Its Variants with Dynamic Weights on CIFAR-100}
  \label{table:gdkd_weights_ablation}
  \resizebox{\linewidth}{!}{
    \begin{threeparttable}
      \begin{tabular}{*{10}{c}}
        \toprule
        \multirow{2}{*}{Method} & $\mathcal{T}$ & ResNet32x4     & ResNet56       & VGG13          & WRN-40-2       & ResNet32x4     & ResNet50       & VGG13          & WRN-40-2       \\
                                & $\mathcal{S}$ & ResNet8x4      & ResNet20       & VGG8           & WRN-16-2       & ShuffleNet-V2  & MobileNet-V2   & MobileNet-V2   & ShuffleNet-V1  \\ \midrule
        GDKD                    &               & \textbf{77.76} & 71.65          & 75.74          & \textbf{76.92} & \textbf{79.27} & 70.55          & 70.35          & 78.05          \\
        GDKD-V1                 &               & 77.57          & 71.48          & 75.67          & 76.62          & 79.08          & 70.65          & 70.00          & 78.03          \\
        GDKD-V2                 &               & 77.70          & \textbf{71.72} & \textbf{75.82} & 76.84          & 79.22          & \textbf{70.92} & \textbf{70.44} & \textbf{78.12} \\
        GDKD-V3                 &               & 77.49          & 71.34          & 75.62          & 76.77          & 79.12          & 70.52          & 69.95          & 78.02          \\\bottomrule
      \end{tabular}
      \begin{tablenotes}
        \item The accuracy is reported over 5 runs on the test set.
      \end{tablenotes}
    \end{threeparttable}
  }
\end{table*}

\begin{table*}[t]
  \centering
  \caption{Results of Logit-based Distillation Methods on CIFAR-100 with Different Temperature}
  \label{table:temp_1_results}
  \resizebox{\linewidth}{!}{
    \begin{threeparttable}
      \begin{tabular}{*{9}{c}}
        \toprule
        \multirow{2}{*}{Temperature} & Teacher                                            & ResNet32x4                         & ResNet56                         & VGG13                              & ResNet32x4                         & ResNet50                           & VGG13                              & WRN-40-2                           \\
                                     & Student                                            & ResNet8x4                          & ResNet20                         & VGG8                               & ShuffleNet-V2                      & MobileNet-V2                       & MobileNet-V2                       & ShuffleNet-V1                      \\ \midrule
        \multirow{4}{*}{T=4}         & KD \cite{hintonDistillingKnowledgeNeural2015}      & 74.68                              & 73.58                            & 74.44                              & 77.25                              & 68.82                              & 68.36                              & 77.06                              \\
                                     & DKD \cite{zhaoDecoupledKnowledgeDistillation2022}  & 76.64                              & 71.15                            & 75.30                              & 78.62                              & 69.88                              & 70.03                              & 77.49                              \\
                                     & DIST \cite{huangKnowledgeDistillationStronger2022} & 77.59                              & 71.27                            & 75.28                              & 78.86                              & 70.01                              & 69.55                              & 77.89                              \\
                                     & \textbf{GDKD}                                      & \textbf{77.76}                     & \textbf{71.65}                   & \textbf{75.74}                     & \textbf{79.27}                     & \textbf{70.55}                     & \textbf{70.35}                     & \textbf{78.05}                     \\\midrule
        \multirow{4}{*}{T=1}         & KD \cite{hintonDistillingKnowledgeNeural2015}      & 73.61 ($\downarrow$ 1.07)          & 70.91 ($\downarrow$ 2.67)        & 72.69 ($\downarrow$ 1.75)          & 74.43 ($\downarrow$ 2.82)          & 66.64 ($\downarrow$ 2.18)          & 66.85 ($\downarrow$ 1.51)          & 75.49 ($\downarrow$ 1.57)          \\
                                     & DKD \cite{zhaoDecoupledKnowledgeDistillation2022}  & 75.26 ($\downarrow$ 1.38)          & 71.40 ($\uparrow$ 0.25)          & 74.13 ($\downarrow$ 1.17)          & 77.42 ($\downarrow$ 1.20)          & 67.00  ($\downarrow$ 2.88)         & 68.64 ($\downarrow$ 1.39)          & 76.54 ($\downarrow$ 0.95)          \\
                                     & DIST \cite{huangKnowledgeDistillationStronger2022} & 74.65 ($\downarrow$ 2.94)          & 71.45 ($\uparrow$ 0.18)          & 72.97 ($\downarrow$ 2.31)          & 75.40 ($\downarrow$ 3.46)          & 67.44  ($\downarrow$ 2.57)         & 67.65 ($\downarrow$ 1.90)          & 75.48 ($\downarrow$ 2.41)          \\
                                     & \textbf{GDKD}                                      & \textbf{77.58 ($\downarrow$ 0.18)} & \textbf{72.09 ($\uparrow$ 0.44)} & \textbf{75.53 ($\downarrow$ 0.21)} & \textbf{78.86 ($\downarrow$ 0.41)} & \textbf{70.14 ($\downarrow$ 0.41)} & \textbf{70.15 ($\downarrow$ 0.20)} & \textbf{78.00 ($\downarrow$ 0.05)} \\\bottomrule
      \end{tabular}
      \begin{tablenotes}
        \item The accuracy is reported over 5 runs on the test set.
      \end{tablenotes}
    \end{threeparttable}
  }
\end{table*}

\subsubsection{Hyperparameter $k$}
\label{sec:ablation_k}
\begin{figure}[t]
  \centering
  \subfloat{
    \includegraphics[width=0.9\linewidth]{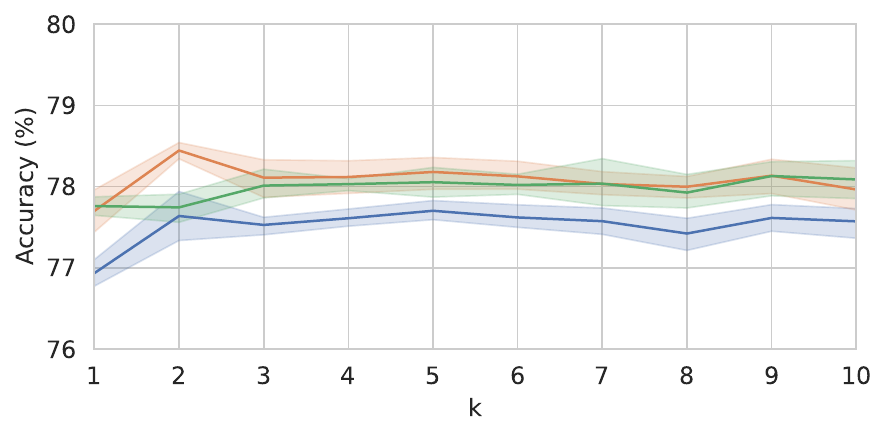}
  }\\
  \subfloat{
    \includegraphics[width=0.9\linewidth]{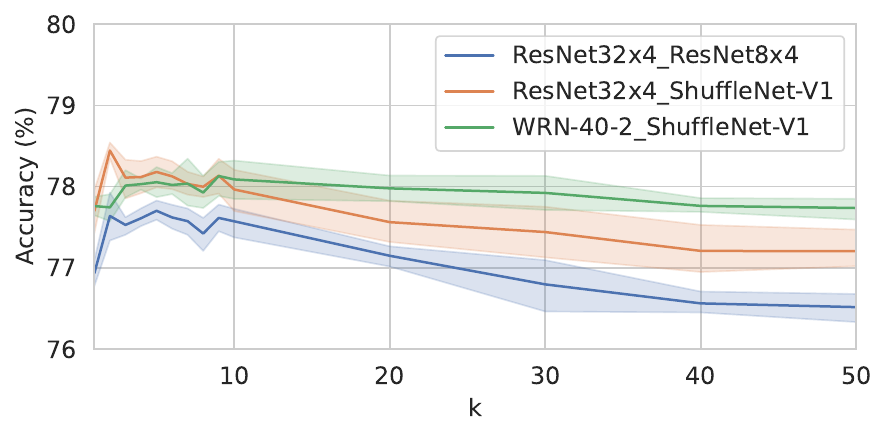}
  }
  \caption{The performance of GDKD with varying $k$ values on the CIFAR-100 test set. The results are bounded by a 95\% confidence interval, derived from 5 independent trials.}
  \label{fig:ablation_k}
\end{figure}

To evaluate the influence of the hyperparameter $k$ in GDKD, we conduct an extensive ablation study on CIFAR-100.
We explore a range of $k$ values, from 1 to 50, including specific increments within 1-10 and intervals of 10 thereafter.
The performance implications of varying $k$ are analyzed using a fixed teacher model (ResNet32x4) with different student models, and a fixed student model (ShuffleNet-V1) with various teacher models.
These results are depicted in Fig.~\ref{fig:ablation_k}.

Our analysis reveals that GDKD's effectiveness is enhanced when $k > 1$. Intriguingly, we observe that performance remains relatively stable with small increases in $k$, but significantly larger values of $k$ lead to a decrease in performance.
This pattern suggests that slightly introducing some small logits into the group with high probabilities is insensitive to the performance since the other majority of small logits can still be effectively distilled. However, including too many small logits in another group might disrupt the balance and relational representation among small logits.
The key lies in maintaining a balance, ensuring logits within each partition have comparable confidence levels for effective distillation.

Consequently, we recommend a heuristic approach for determining $k$.
This involves analyzing the predictive distribution of the pre-trained teacher model to identify an optimal $k$ value that encompasses classes with significant soft prediction values.
It is important to note that the ideal value of $k$ is contingent on the interplay between the teacher model and dataset characteristics.
For instance, factors such as the teacher model’s architecture and capacity, along with the training approach (e.g., data augmentation techniques), are pivotal in shaping the predictive distribution and consequently, the appropriate selection of $k$.
Meanwhile, in datasets like CIFAR-100, certain classes (e.g., aquatic animals) may exhibit inherent overlapping features, implicitly influencing the suitable $k$ value for a strong teacher model.

\subsubsection{Loss Components}
\label{sec:ablation_gdkd_terms}
In the GDKD algorithm, the KD loss is hierarchically split into three distinct components: highKD ($\text{KL}(\bm{b}^\mathcal{T}||\bm{b}^\mathcal{S})$), lowKD-topk ($\text{KL}(\bm{p}_\mathbb{T}^\mathcal{T}||\bm{p}\mathbb{T}^\mathcal{S})$), and lowKD-other ($\text{KL}(\bm{p}_{\setminus \mathbb{T}}^\mathcal{T}||\bm{p}_{\setminus \mathbb{T}}^\mathcal{S})$).
These components are assigned weights $w_0$, $w_1$, and $w_2$ respectively. To understand the individual and collective impact of these components, we carry out an ablation study on the CIFAR-100 dataset, examining various combinations of these terms.
The results are comprehensively presented in Table~\ref{table:gdkd_terms_ablation}.

Our findings indicate that the lowKD-other component is an essential driver of the performance enhancements in GDKD.
When used alone with its weight $w_2$, this component already outperforms the traditional knowledge distillation method.
This observation aligns with our discussion in Section~\ref{sec:empirical_analysis} about the \emph{dark knowledge} present in the relationships among these small logits.
Both KD and DKD tend to overlook this suppression of knowledge due to their coupling of small and large logits.
GDKD, by contrast, effectively isolates small logits into a separate group, enhancing their relational structure via recalibrated softmax predictions.

Additionally, the study shows that the other two components, highKD and lowKD-topk, also capture additional knowledge from the teacher model.
The integration of all three components in GDKD culminates in its superior overall performance, leveraging the full spectrum of knowledge available from the teacher model.

\subsubsection{Loss Weights}
\label{sec:ablation_gdkd_weights}
In GDKD, we apply fixed loss weights $w_0$, $w_1$, and $w_2$ to the different distillation components.
However, considering the potential benefits of a more adaptive approach, we explore the implementation of dynamic scaling factors $m_1$ and $m_2$, multiplied by the loss coefficients in original KD loss, specifically $b_{\mathbb{T}}^\mathcal{T}$ and $b_{\setminus \mathbb{T}}^\mathcal{T}$.
This leads to the development of three GDKD variants (GDKD-V1, GDKD-V2, and GDKD-V3), each utilizing distinct modifications to their loss functions to integrate dynamic scaling:
\begin{equation}
  \begin{aligned}
    \mathcal{L}_{\text{V1}} & =\mathcal{L}_{\text{highKD}}  + m_1 \cdot b_\mathbb{T}^\mathcal{T} \mathcal{L}_{\text{lowKD-topk}} + m_2 \cdot b_{\setminus \mathbb{T}}^\mathcal{T} \mathcal{L}_{\text{lowKD-other}}, \\
    \mathcal{L}_{\text{V2}} & = \mathcal{L}_{\text{highKD}}  + w_1 \mathcal{L}_{\text{lowKD-topk}} + m_2 \cdot b_{\setminus \mathbb{T}}^\mathcal{T} \mathcal{L}_{\text{lowKD-other}},                               \\
    \mathcal{L}_{\text{V3}} & = \mathcal{L}_{\text{highKD}}  + m_1 \cdot b_\mathbb{T}^\mathcal{T} \mathcal{L}_{\text{lowKD-topk}} + w_2 \mathcal{L}_{\text{lowKD-other}}.
  \end{aligned}
  \label{eq:gdkd_autow_loss}
\end{equation}

To assess the impact of static versus dynamic weighting schemes, we conduct comparative experiments on the CIFAR-100 dataset.
The dynamic scaling factors $m_1$ and $m_2$ in these variants are calculated such that the mean values of $m_1 \cdot b_{\mathbb{T}}^{\mathcal{T}}$ and $m_2 \cdot b_{\setminus \mathbb{T}}^{\mathcal{T}}$ approximately match the fixed weights $w_1$ and $w_2$.
The comparative results, detailed in Table~\ref{table:gdkd_weights_ablation}, reveal that GDKD-V2 achieves marginal performance gains only in certain tasks, and both GDKD-V1 and GDKD-V3 are generally underperformed in comparison with the standard GDKD configuration.

These findings suggest that dynamic adjustment of the loss weights might not significantly enhance the distillation process.
Therefore, we choose to proceed with the simpler and computationally efficient fixed-weights approach in GDKD, which also facilitates a faster training process.

\subsection{Extensive Results}

\begin{table*}[b]
  \centering
  \caption{Results of DKD and GDKD-top1 on CIFAR-100 with Different Data Augmentation Strategies on Teacher Models}
  \label{table:dkd_vs_gdkd_top1}
  \resizebox{\linewidth}{!}{
    \begin{threeparttable}
      \begin{tabular}{*{10}{c}}
        \toprule
        \multirow{2}{*}{Teacher Type}                                            & Teacher                                           & ResNet32x4     & ResNet56       & VGG13          & WRN-40-2       & ResNet32x4     & ResNet50       & VGG13          & WRN-40-2       \\
                                                                                 & Student                                           & ResNet8x4      & ResNet20       & VGG8           & WRN-16-2       & ShuffleNet-V2  & MobileNet-V2   & MobileNet-V2   & ShuffleNet-V1  \\ \midrule
        \multirow{3}{*}{Standard}                                                & DKD \cite{zhaoDecoupledKnowledgeDistillation2022} & \textbf{76.30} & \textbf{71.60} & \textbf{74.68} & \textbf{75.80} & \textbf{77.00} & \textbf{70.55} & \textbf{69.66} & \textbf{76.70} \\
                                                                                 & GDKD-top1                                         & 76.26          & 71.46          & 74.60          & 75.71          & 76.91          & 70.42          & 69.59          & 76.48          \\ 
                                                                                 & $\Delta$                                          & -0.04          & -0.14          & -0.08          & -0.09          & -0.09          & -0.13          & -0.07          & -0.22          \\ \midrule
        \multirow{3}{*}{AutoAug \cite{cubukAutoAugmentLearningAugmentation2019}} & DKD \cite{zhaoDecoupledKnowledgeDistillation2022} & 76.64          & 71.15          & 75.30          & 76.04          & 78.62          & 69.88          & \textbf{70.03} & 77.49          \\
                                                                                 & GDKD-top1                                         & \textbf{76.93} & \textbf{71.39} & \textbf{75.42} & \textbf{76.15} & \textbf{78.71} & \textbf{70.31} & 69.96          & \textbf{77.76} \\
                                                                                 & $\Delta$                                          & +0.29          & +0.24          & +0.12          & +0.11          & +0.09          & +0.43          & -0.07          & +0.27          \\ \bottomrule
      \end{tabular}
      \begin{tablenotes}
        \item The accuracy is reported over 5 runs on the test set and $\Delta$ represents the accuracy difference between GDKD-top1 and DKD.
      \end{tablenotes}
    \end{threeparttable}

  }
\end{table*}

\begin{figure}[t!]
  \centering
  \subfloat{
    \includegraphics[width=0.49\linewidth]{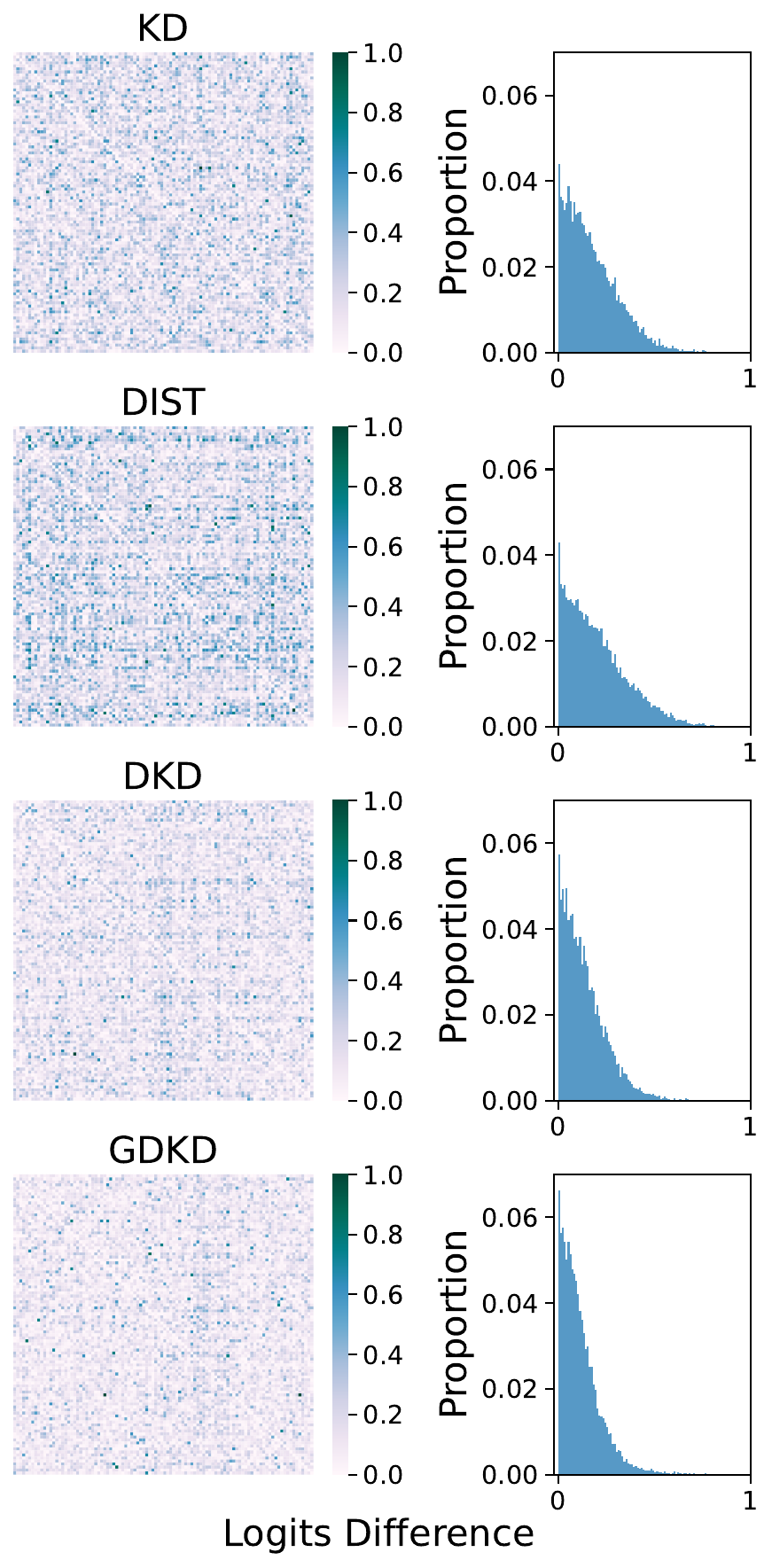}
  }
  \subfloat{
    \includegraphics[width=0.49\linewidth]{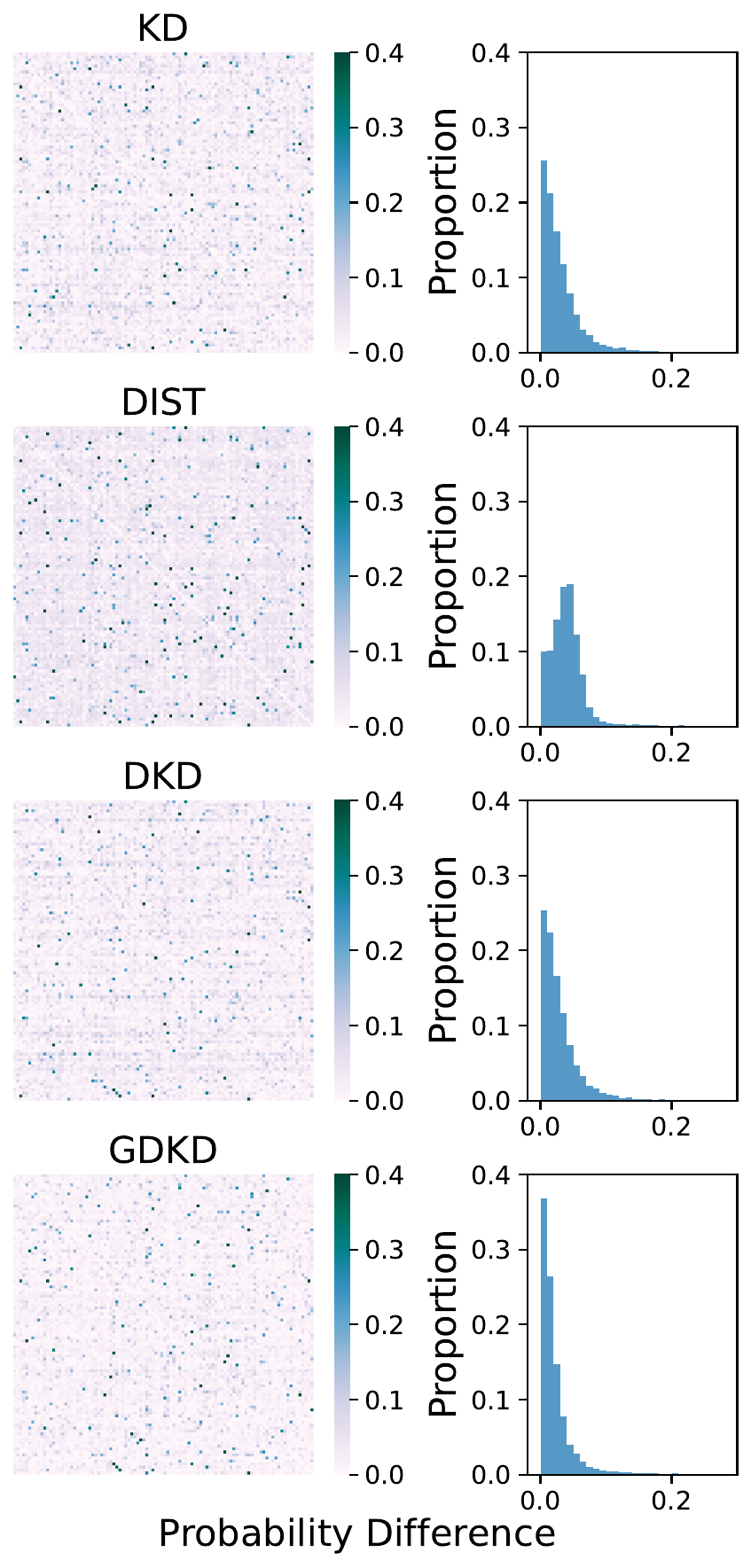}
  }
  \caption{The logit and probability ($T=4$) differences between the teacher model ResNet32x4 and the student model ShuffleNet-V1 with different logit-based knowledge distillation methods on the CIFAR-100 test set. We visualize the average discrepancy over each class and report the difference distribution. The differences on the target label (i.e., the diagonal values) are ignored.}
  \label{fig:logit_prob_diff}
\end{figure}

\subsubsection{Temperature Setting $T=1$}
\label{sec:ablation_temp}
To further explore GDKD's capability in transferring knowledge from small logits, we conduct a series of experiments on CIFAR-100 using a softmax temperature of $T = 1$.
At this temperature, the soft prediction values of small logits become significantly small after softmax, challenging the knowledge extraction from them.
The performance results, presented in Table~\ref{table:temp_1_results}, reveal that while DKD and other logit-based distillation methods face notable declines in performance under these conditions, GDKD maintains robustness with minimal performance degradation. Additionally, in the case where ResNet56 serves as the teacher model and ResNet20 as the student, GDKD not only sustains its performance but also achieves the most significant improvements compared to other KD methods.
This highlights GDKD's exceptional ability to effectively harness and transfer the nuanced knowledge contained within the raw logits, where most logits have extremely minimal probabilities.

\subsubsection{Impact Analysis of GDKD}
To better understand how GDKD influences student models, we investigate the average discrepancies in logits and soft predictions between teacher-student pairs.
This analysis, depicted in Fig.~\ref{fig:logit_prob_diff}, specifically examines how GDKD affects the alignment of predictions for non-target classes.
The results indicate a marked improvement by GDKD in reducing differences in both logits and corresponding soft predictions. This enhanced alignment is primarily due to GDKD's focused distillation strategy, which places greater emphasis on learning from small logits.
By effectively decoupling and concentrating on these small logits, GDKD facilitates a more precise and balanced knowledge transfer from teacher to student models.

\subsubsection{DKD \textit{vs.} GDKD-top1}
While DKD can be seen as a variant of GDKD employing a top-1 partitioning strategy (GDKD-top1), there are subtle yet significant differences in their partitioning methodologies. DKD primarily focuses on the target label for partitioning, whereas GDKD-top1 emphasizes the highest confidence label identified by the teacher model. To explore the practical implications of these distinct strategies, we conduct a comparative analysis, the results of which are detailed in Table~\ref{table:dkd_vs_gdkd_top1}. Our experiments reveal that in cases where teacher models exhibit multimodal predictions, GDKD-top1 generally demonstrates superior performance compared to DKD. This suggests that the top-logit approach of GDKD-top1 is more effective in leveraging the diverse knowledge present in such scenarios.
However, in situations where the teacher models exhibit unimodal predictions, possibly due to overfitting, GDKD-top1 is slightly outperformed by DKD.
This indicates that the target label-based approach of DKD may offer some extra knowledge.

These observations suggest that, while the target label might provide some benefits in certain cases, its advantages are less pronounced in scenarios with multimodal predictions from the teacher, where the top-1 partitioning strategy in GDKD-top1 seems to be more effective.


\section{Conclusion}
\label{sec:conclusion}

This research pushes the envelope of logit-based knowledge distillation, delving into the deep mechanics of logit decoupling and its impact on the distillation process.
Central to our study was the introduction of Generalized Decoupled Knowledge Distillation (GDKD), providing a fine-grained and flexible approach for logit-based knowledge distillation in the decoupling paradigm.

Under the GDKD framework, our analysis summarized two key factors for the effectiveness of logit decoupling in previous studies.
This included an in-depth examination of teachers' prediction distributions and their linkage to the gradients of different decoupled loss terms in GDKD, thereby providing a novel perspective on the principles of decoupled knowledge distillation. Derived from these insights, the GDKD algorithm was developed. GDKD enhances the knowledge expression and extraction among small logits, unlocking the potential of logit-based knowledge. Verified across a range of diverse tasks, the proposed algorithm demonstrated superior performance over state-of-the-art logit-based methods, including DKD, and various feature-based approaches, indicating the effectiveness of GDKD, especially in scenarios where teacher models produce multimodal predictive distributions.

Despite these promising developments, our findings also point to areas where further improvements are needed.
The determination of weight coefficients in the GDKD loss (specifically $w_1$ and $w_2$) currently lacks a theoretical underpinning, relying instead on empirical practice (see Appendix~\ref{sec:train_impl}). Nevertheless, the total number of hyperparameters for tuning is similar to that of DKD. Since the experiences in Sec.~\ref{sec:ablation_temp} demonstrated the effectiveness of GDKD on raw logits, the distillation temperature can be exempted. Additionally, the value of $k$ can be easily set by examining the knee point of the teacher's predictive distribution and is less sensitive within a rational range as shown in Sec.~\ref{sec:ablation_k}.
Additionally, the simple two-partition decoupling in our experiments could potentially limit the adaptability and overall efficacy of the distillation process.
Future research will aim to develop an adaptive mechanism for diverse partitioning and establish a theoretical framework for determining the optimal weights of decoupled losses.
Such advancements will undoubtedly enhance the robustness and applicability of GDKD.

  {
    \appendices

    \section{Implementation Details}
    \label{sec:train_impl}

    Our implementation is mainly developed on the code of DKD\footnote{\url{https://github.com/megvii-research/mdistiller}} \cite{zhaoDecoupledKnowledgeDistillation2022}, which implements a plethora of KD methods with a standard training procedure.

    \textbf{Evaluation Metrics:} We follow the evaluation metrics used in DKD \cite{zhaoDecoupledKnowledgeDistillation2022}, which records the best accuracies among epochs and reports the average of best accuracies among multiple independent runs.

    \textbf{CIFAR-100}: The batch size is 64 with 240 epochs. The optimizer is SGD with weight decay 5e-4 and momentum 0.9. For the learning rate, ShuffleNet \cite{zhangShuffleNetExtremelyEfficient2017} and MobileNet-V2 \cite{sandlerMobileNetV2InvertedResiduals2019} use 0.01, other models (VGG \cite{simonyanVeryDeepConvolutional2015}, ResNet \cite{heDeepResidualLearning2015} and WRN \cite{zagoruykoWideResidualNetworks2017}) use 0.05. The learning rate is decayed by factor 10 at 150, 180, and 210 epochs. For logit-based methods, a 20-epoch linear warmup of distillation loss in DKD and GDKD is utilized for stable training. The weight for the student's cross-entropy loss is fixed to 1. And the temperature $T$ is uniformly set to 4, except for LS, which utilizes $T=2$ \cite{sunLogitStandardizationKnowledge2024}. For the hyperparameters of loss weights in GDKD, i.e., $w_0,w_1,w_2"$, we set $w_0=1$, and reference the value of $\beta$ in DKD to set $w_2$, which are heuristically set by the gap $z_t - z_{\max}$ (see Appendix A.3 in DKD\cite{zhaoDecoupledKnowledgeDistillation2022}). For the value of $w_1$, we sweep from 1 to half of $w_2$. Table~\ref{table:gdkd_loss_weights} summarizes the loss weights of GDKD in our experiments on CIFAR-100, along with the scale coefficients $m1,m2$ used in Section~\ref{sec:ablation_gdkd_terms}. For experiments on Section~\ref{sec:empirical_analysis}, teacher models are pre-trained on CIFAR-100 with standard data augmentation. For CIFAR-100 experiments on Section~\ref{sec:exp}, teacher models are pre-trained with AutoAugment \cite{cubukAutoAugmentLearningAugmentation2019}.

    \begin{table}[t]
      \centering
      \caption{The Loss Weights Used in GDKD on CIFAR-100}
      \label{table:gdkd_loss_weights}
      \begin{tabular}{*{6}{c}}
        \toprule
        Teacher    & Student       & $w_1$ & $w_2$ & $m_1$ & $m_2$ \\ \midrule
        ResNet56   & ResNet20      & 1     & 1     & 3     & 2     \\
        ResNet110  & ResNet32      & 1     & 1     & -     & -     \\
        WRN-40-2   & ShuffleNet-V1 & 2     & 6     & 6     & 10    \\
        WRN-40-2   & WRN-16-2      & 2     & 6     & 6     & 10    \\
        WRN-40-2   & WRN-40-1      & 2     & 6     & -     & -     \\
        VGG13      & VGG8          & 1     & 6     & 3     & 10    \\
        VGG13      & MobileNet-V2  & 1     & 6     & 3     & 10    \\
        ResNet50   & MobileNet-V2  & 1     & 8     & 2     & 16    \\
        ResNet32x4 & ResNet8x4     & 2     & 8     & 6     & 14    \\
        ResNet32x4 & ShuffleNet-V1 & 1     & 8     & -     & -     \\
        ResNet32x4 & ShuffleNet-V2 & 1     & 8     & 3     & 14    \\\bottomrule
      \end{tabular}
    \end{table}

    \textbf{ImageNet:} Every model is trained with 100 epochs with batch size 512. The optimizer is SGD with weight decay 1e-4 and momentum 0.9. The learning rate is 0.1 and decayed by factor 10 at 30, 60, and 90 epochs. The temperature $T$ is set to 1, except that LS utilizes $T=2$ \cite{sunLogitStandardizationKnowledge2024}. The distillation warmup is 1 epoch for DKD and GDKD. For loss weights of GDKD, we also reference the DKD settings: when using ResNet34 as the teacher, set $w_0=0.5, w_1=w_2=1.0$; when using ResNet50 as the teacher, set $w_0=0.5, w_1=w_2=2.0$.

    \textbf{Tiny-Imagenet \& CUB-200-2011:} After obtaining the pre-trained models on ImageNet, the teacher model is firstly fine-tuned on the training set of Tiny-Imagenet or CUB-200-2011 for 100 epochs with a batch size of 64. The optimizer is SGD with weight decay 1e-3 and momentum 0.9. On Tiny-ImageNet, the learning rate is 1e-3 and decayed by factor 10 at 30 and 60 epochs. On CUB-200-2011, the learning rate is 2.5e-3 and decayed by factor 10 at 40 and 80 epochs; the strong data augmentation on the teacher models includes RandomResizedCrop, RandomAugmentation \cite{cubukRandAugmentPracticalAutomated2020} and RandomErasing \cite{zhongRandomErasingData2020}. For logit-based methods, the temperature $T$ is set to 1. The distillation warmup epochs and the loss weights in DKD and GDKD remain the same values as the ImageNet experiments.

    \textbf{Cityscapes}: The implementation on segmentation tasks is based on DIST\footnote{\url{https://github.com/hunto/DIST_KD}} \cite{huangKnowledgeDistillationStronger2022}. The student model is trained with 40k iterations, with 16 images per iteration. The learning rate is 0.02, exponentially decayed by power factor 2.0 after each iteration. And the optimizer is SGD with weight decay 5e-4 and momentum 0.9. The temperature $T$ is set to 1 for logit-based KD methods. Since the predictions of both the teacher models and the student models are downsampled in a smaller size compared to the input image size, to avoid the information loss from the downsampling of the target label in DKD and align with other logit-based methods, we upscale the predictions to the original image size instead, then calculate the average per-pixel distillation loss for all logit-based KD methods. The loss weights of DKD are $\alpha=1, \beta=0.25$, and the loss weights of GDKD are $w_0=1,w_1=0.5,w_2=0.25$.

    \section{Proof of Eq.~\eqref{eq:kd_loss_gdkd}}
    \label{appendix:gdkd_loss_proof}

    \begin{proof}
      By definition, we have $ p_i = \frac{\exp(z_i)}{\sum_j^C \exp(z_j)}, b_{\mathbb{T}} = \frac{\sum_{i \in \mathbb{T}} \exp(z_i)}{\sum_j^C \exp(z_j)}, p_{i,\mathbb{T}} = \frac{\exp(z_i)}{\sum_{j \in \mathbb{T}} \exp(z_j) }$, where $p_i = b_{\mathbb{T}} \cdot p_{i,\mathbb{T}}$. And we can rewrite the KD loss in Eq.~\eqref{eq:kd_loss} as:
      \begin{equation}
        \begin{aligned}
          \mathcal{L}_\text{KD} & = \text{KL}(p^\mathcal{T}||p^\mathcal{S})                                                                                                                                                   \\
                                & = \sum_i^C p_i^\mathcal{T} \log \frac{p_i^\mathcal{T}}{p_i^\mathcal{S}}                                                                                                                     \\
                                & = \sum_{i \in \mathbb{T}} p_i^\mathcal{T} \log \frac{p_i^\mathcal{T}}{p_i^\mathcal{S}} + \sum_{i \in {\setminus \mathbb{T}} } p_i^\mathcal{T} \log \frac{p_i^\mathcal{T}}{p_i^\mathcal{S}}.
        \end{aligned}
      \end{equation}
      Then for $\sum_{i \in \mathbb{T}} p_i^\mathcal{T} \log \frac{p_i^\mathcal{T}}{p_i^\mathcal{S}}$, we have:
      \begin{equation}
        \begin{aligned}
            & \sum_{i \in \mathbb{T}} p_i^\mathcal{T} \log \frac{p_i^\mathcal{T}}{p_i^\mathcal{S}}                                                                                                                                                                                                                                                             \\
          = & \sum_{i \in \mathbb{T}}  b_{\mathbb{T}}^\mathcal{T}  p_{i,\mathbb{T}}^\mathcal{T} \log \frac{ b_{\mathbb{T}}^\mathcal{T}  p_{i,\mathbb{T}}^\mathcal{T}}{ b_{\mathbb{T}}^\mathcal{S}  p_{i,\mathbb{T}}^\mathcal{S}}                                                                                                                               \\
          = & b_{\mathbb{T}}^\mathcal{T} \sum_{i \in \mathbb{T}}    p_{i,\mathbb{T}}^\mathcal{T} \left[ \log \frac{ b_{\mathbb{T}}^\mathcal{T} }{ b_{\mathbb{T}}^\mathcal{S}} + \log \frac{p_{i,\mathbb{T}}^\mathcal{T}}{p_{i,\mathbb{T}}^\mathcal{S}} \right]                                                                                                 \\
          = & b_{\mathbb{T}}^\mathcal{T} \log \frac{ b_{\mathbb{T}}^\mathcal{T} }{ b_{\mathbb{T}}^\mathcal{S}}  \sum_{i \in \mathbb{T}}   \left[ p_{i,\mathbb{T}}^\mathcal{T} \right] + b_{\mathbb{T}}^\mathcal{T} \sum_{i \in \mathbb{T}} \left[ p_{i,\mathbb{T}}^\mathcal{T} \log \frac{p_{i,\mathbb{T}}^\mathcal{T}}{p_{i,\mathbb{T}}^\mathcal{S}}  \right] \\
          = & b_{\mathbb{T}}^\mathcal{T} \log \frac{ b_{\mathbb{T}}^\mathcal{T} }{ b_{\mathbb{T}}^\mathcal{S}}  + b_{\mathbb{T}}^\mathcal{T} \text{KL}(\bm{p}_{\mathbb{T}}^\mathcal{T} ||  \bm{p}_{\mathbb{T}}^\mathcal{S}).
        \end{aligned}
      \end{equation}
      Similarly, for $\sum_{i \in {\setminus \mathbb{T}} } p_i^\mathcal{T} \log \frac{p_i^\mathcal{T}}{p_i^\mathcal{S}}$, we have
      \begin{equation}
        \sum_{i \in {\setminus \mathbb{T}} } p_i^\mathcal{T} \log \frac{p_i^\mathcal{T}}{p_i^\mathcal{S}} = b_{\setminus \mathbb{T}}^\mathcal{T} \log \frac{ b_{ \setminus \mathbb{T}}^\mathcal{T} }{ b_{\setminus \mathbb{T}}^\mathcal{S}}  + b_{\setminus \mathbb{T}}^\mathcal{T} \text{KL}(\bm{p}_{\setminus \mathbb{T}}^\mathcal{T} ||  \bm{p}_{\setminus \mathbb{T}}^\mathcal{S}).
      \end{equation}
      Then combining the above two equations, we can further rewrite the KD loss by:
      \begin{equation}
        \begin{aligned}
          \mathcal{L}_\text{KD} = &
          b_{\mathbb{T}}^\mathcal{T} \log \frac{ b_{\mathbb{T}}^\mathcal{T} }{ b_{\mathbb{T}}^\mathcal{S}} + b_{\setminus \mathbb{T}}^\mathcal{T} \log \frac{ b_{ \setminus \mathbb{T}}^\mathcal{T} }{ b_{\setminus \mathbb{T}}^\mathcal{S}}                                                                                                 \\
                                  & +b_{\mathbb{T}}^\mathcal{T} \text{KL}(\bm{p}_{\mathbb{T}}^\mathcal{T} ||  \bm{p}_{\mathbb{T}}^\mathcal{S}) + b_{\setminus \mathbb{T}}^\mathcal{T} \text{KL}(\bm{p}_{\setminus \mathbb{T}}^\mathcal{T} || \bm{p}_{\setminus \mathbb{T}}^\mathcal{S})                                                      \\
          =                       & \text{KL} (\bm{b}^\mathcal{T} || \bm{b}^\mathcal{S}) +b_{\mathbb{T}}^\mathcal{T} \text{KL}(\bm{p}_{\mathbb{T}}^\mathcal{T} || \bm{p}_{\mathbb{T}}^\mathcal{S}) + b_{\setminus \mathbb{T}}^\mathcal{T} \text{KL}(\bm{p}_{\setminus \mathbb{T}}^\mathcal{T} || \bm{p}_{\setminus \mathbb{T}}^\mathcal{S}).
        \end{aligned}
      \end{equation}
    \end{proof}

    \section{Supplementary Experiments}

    \subsection{GDKD on Transformer Models}

    \begin{table}[h]
      \centering
      \caption{Results of Different Distillation Methods on the ImageNet Validation Set with Vision Transformer Teachers}
      \label{table:gdkd3_vit}
      \begin{threeparttable}
        \begin{tabular}{*{3}{c}}
          \toprule
          Type                    & Top-1 Accuracy & Top-5 Accuracy \\ \midrule
          Teacher: Swin-B         & 83.36          & 96.56          \\
          Student: WideResNet50-2 & 78.47          & 94.09          \\ \midrule
          KD                      & 78.70          & 94.27          \\
          DKD                     & 79.56          & 94.76          \\
          DIST                    & 78.45          & 94.21          \\
          GDKD3                   & \textbf{79.81} & \textbf{94.78} \\ \bottomrule
        \end{tabular}
        \begin{tablenotes}
          \item The Top-1 and Top-5 accuracy are collected from the average of 3 independent runs.
        \end{tablenotes}
      \end{threeparttable}
    \end{table}

    To further validate the generalizability of our methodology, we extend the experiments to Transformer-based architectures. Specifically, we adopt Swin-B \cite{liu2021swin} as the teacher model and WideResNet50-2 \cite{zagoruykoWideResidualNetworks2017} as the student model.
    Upon examining the logit distribution of the teacher, we observe that its multimodal predictive distribution differs slightly from that of CNN models: the softmax probability of the top-1 logit is substantially higher than the cumulative probability of the remaining top-$k$ logits.
    This imbalance may hinder effective knowledge representation among top-$k$ logits under the original GDKD partitioning scheme.

    To address this, we propose a three-partition variant of GDKD, denoted as GDKD3, following the formulation in Eq.~\eqref{eq:gdkd_loss_n}.
    The logits are divided into three groups: (1) the top-1 logit, (2) the top-2 to top-$k$ logits (we empirically set $k = 5$), and (3) all remaining logits.
    The loss function maintains the same number of hyperparameters as GDKD, with $w_0$, $w_1$, and $w_2$ respectively corresponding to the top-level KD loss, the second group, and the third group.
    In our implementation, all weights are set to 1.

    We compare GDKD3 against other logit-based knowledge distillation methods on the ImageNet dataset.
    Both teacher and student models are trained using the official PyTorch recipes\footnote{\url{https://docs.pytorch.org/vision/main/models.html\#classification}}.
    To ensure the integrity of soft logit knowledge, we disable label smoothing during teacher training and omit distillation in the first epoch for DKD and GDKD3.

    As evidenced by the results in Table~\ref{table:gdkd3_vit}, GDKD3 achieves the best performance among all evaluated methods.
    This demonstrates that our proposed GDKD framework can be effectively applied to Transformer-based architectures, further supporting the robustness and general applicability of the proposed approach.

    \subsection{Extended Ablation Study of k}
    \begin{table*}[t]
      \centering
      \caption{Results of GDKD with Logit Standardization on the CIFAR-100 Test Set}
      \label{table:gdkd_with_ls}
      \begin{threeparttable}
        \begin{tabular}{*{10}{c}}
          \toprule
          \multirow{2}{*}{Method} & $\mathcal{T}$ & ResNet32x4     & ResNet56       & VGG13          & WRN-40-2       & ResNet32x4     & ResNet50       & VGG13          & WRN-40-2       \\
                                  & $\mathcal{S}$ & ResNet8x4      & ResNet20       & VGG8           & WRN-40-1       & ShuffleNet-V1  & MobileNet-V2   & MobileNet-V2   & ShuffleNet-V1  \\ \midrule
          GDKD                    &               & 77.76          & 71.65          & 75.74          & \textbf{75.75} & 78.21          & 70.55          & 70.35          & \textbf{78.05} \\
          GDKD w/ LS              &               & \textbf{78.48} & \textbf{71.67} & \textbf{75.81} & 75.70          & \textbf{78.57} & \textbf{71.17} & \textbf{70.90} & 78.01          \\
          $\Delta$                &               & +0.72          & +0.02          & +0.07          & -0.05          & +0.36          & +0.62          & +0.55          & -0.04          \\\bottomrule
        \end{tabular}
        \begin{tablenotes}
          \item $\Delta$ represents the performance improvement of GDKD from DKD. All results are averaged over 5 independent runs.
        \end{tablenotes}
      \end{threeparttable}
    \end{table*}

    To further examine the influence of the hyperparameter $k$ in GDKD across different datasets, we perform an additional ablation study under the transfer learning setting on CUB-200-2011.
    Specifically, we evaluate the same two teacher-student model pairs described in Section~\ref{sec:transfer_learning_results}, varying the value of $k$.
    As illustrated in Fig.~\ref{fig:ablation_k_TL}, the results are consistent with those observed on CIFAR-100 in Section~\ref{sec:ablation_k}.
    The performance of GDKD remains relatively stable within a reasonable range of $k$, but degrades when $k$ is set too small or too large.

    \begin{figure}
      \centering
      \includegraphics[width=\linewidth]{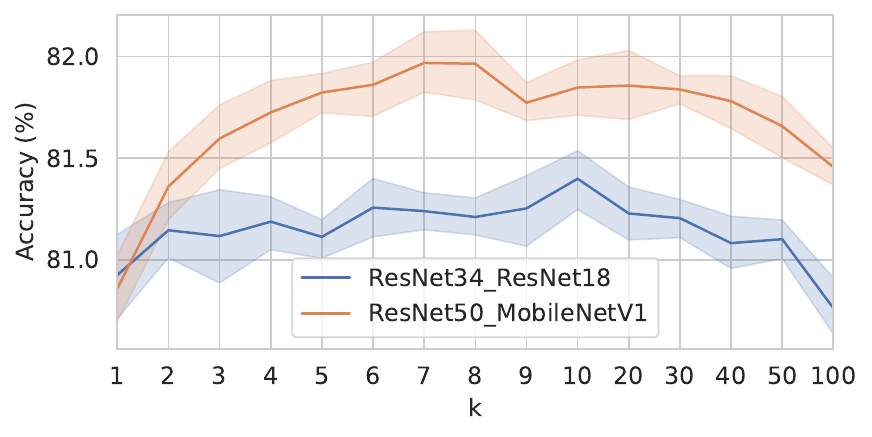}
      \caption{The performance of GDKD with varying $k$ values on CUB-200-2011 test set. The results are bounded by a 95\% confidence interval, derived from 9 independent trials.}
      \label{fig:ablation_k_TL}
    \end{figure}

    \subsection{Integration with Other Logit-based Distillation Methods}
    To further examine the compatibility of GDKD with other logit-based distillation techniques, we integrate GDKD with the recently proposed Logit Standardization (LS) method~\cite{sunLogitStandardizationKnowledge2024}.
    Specifically, a z-score normalization is applied to the logits prior to computing the GDKD loss, and the loss is uniformly scaled by an additional factor of 9.0, following the configuration recommended in~\cite{sunLogitStandardizationKnowledge2024}.
    Experiments are conducted on the CIFAR-100 dataset with the temperature set to $T=4.0$.
    As summarized in Table~\ref{table:gdkd_with_ls}, the results show that combining GDKD with LS further enhances accuracy for several teacher-student pairs, particularly those with heterogeneous architectures.
    This observation suggests that GDKD and LS address complementary aspects of the distillation process, such that additional performance gains can be achieved by jointly employing both methods.
  }

\bibliographystyle{IEEEtran}
\bibliography{ref}







\begin{IEEEbiography}[{\includegraphics[height=1.25in,clip,keepaspectratio]{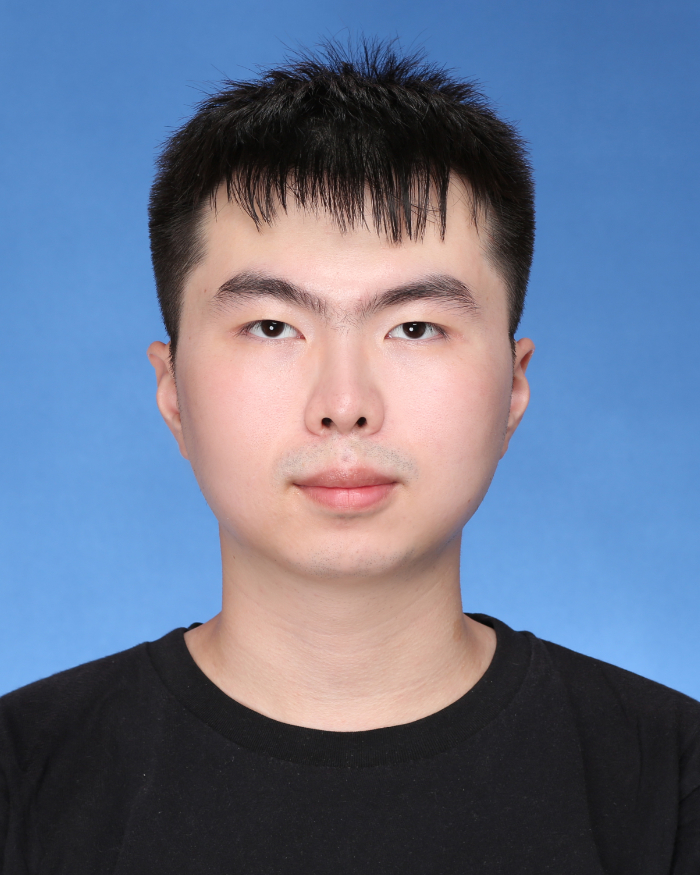}}]
    {Bowen~Zheng} is currently pursuing the PhD degree at the Department of Data Science and Artificial Intelligence, The Hong Kong Polytechnic University. He received the B.Eng. degree and the M.Eng. degree from the Southern University of Science and Technology.
    His research interests include neural architecture search and evolutionary reinforcement learning.
\end{IEEEbiography}

\begin{IEEEbiography}[{\includegraphics[width=1in,height=1.25in,clip,keepaspectratio]{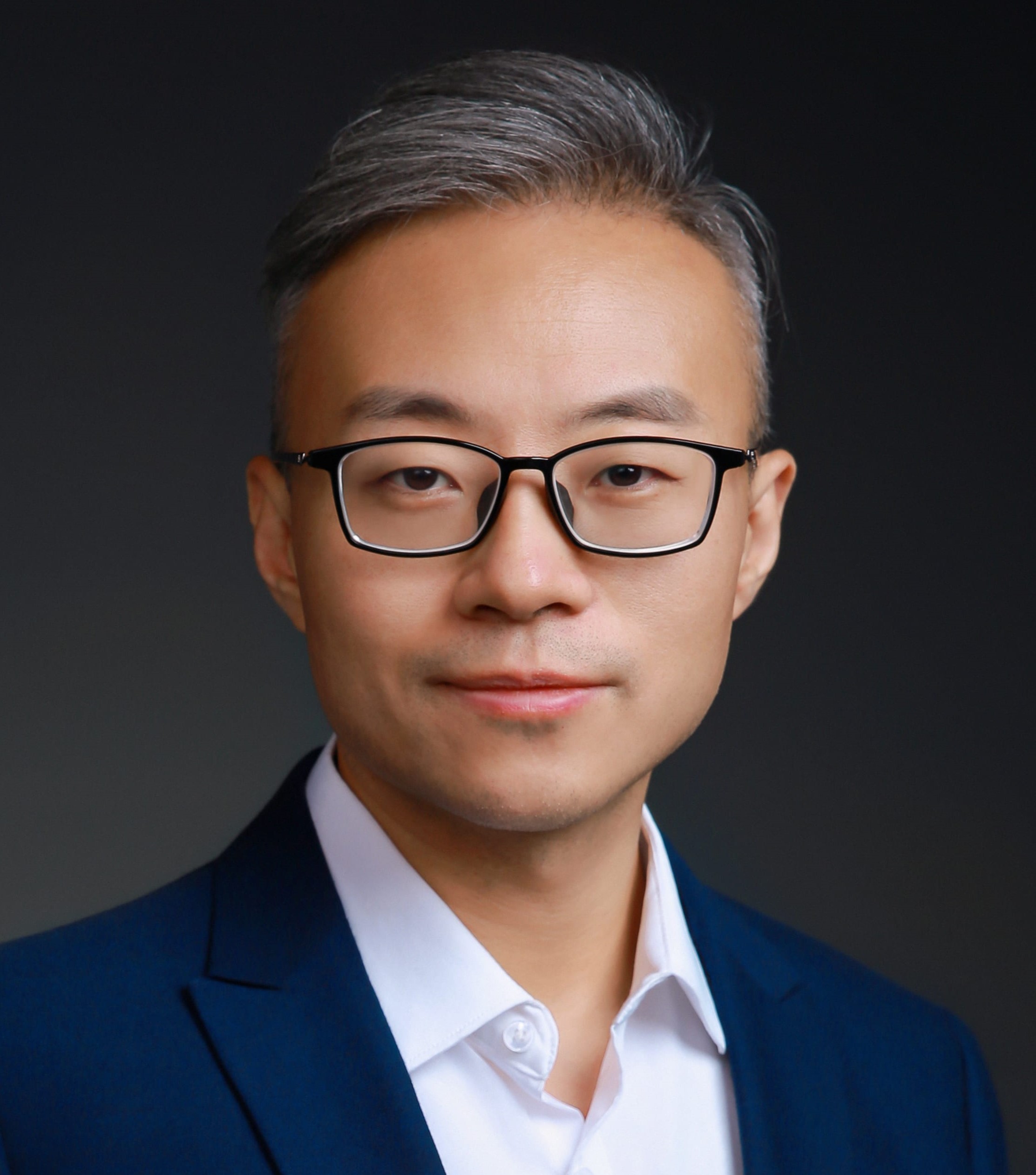}}]{Ran Cheng} (Senior Member, IEEE)
received the B.Sc. degree from the Northeastern University, Shenyang, China, in 2010, and the Ph.D. degree from the University of Surrey, Guildford, U.K., in 2016. 
He is currently an Associate Professor with the Department of Data Science and Artificial Intelligence, and the Department of Computing, The Hong Kong Polytechnic University, Hong Kong SAR, China. 
He is a recipient of the IEEE Transactions on Evolutionary Computation Outstanding Paper Award (2018 and 2021), the IEEE Computational Intelligence Society Outstanding Ph.D. Dissertation Award (2019), the IEEE Computational Intelligence Magazine Outstanding Paper Award (2020), and the IEEE Computational Intelligence Society Early Career Award (2025). 
He is the Founding Chair of the IEEE Computational Intelligence Society Shenzhen Chapter. 
He is an Associate Editor of IEEE Transactions on Evolutionary Computation, IEEE Transactions on Artificial Intelligence, IEEE Transactions on Emerging Topics in Computational Intelligence, and IEEE Transactions on Cognitive and Developmental Systems.
\end{IEEEbiography}


\end{document}